\documentclass{article}

    \usepackage[final,nonatbib]{neurips_2021}

\usepackage[utf8]{inputenc} % allow utf-8 input
\usepackage[T1]{fontenc}    % use 8-bit T1 fonts
\usepackage{hyperref}       % hyperlinks
\usepackage{url}            % simple URL typesetting
\usepackage{booktabs}       % professional-quality tables
\usepackage{amsfonts}       % blackboard math symbols
\usepackage{nicefrac}       % compact symbols for 1/2, etc.
\usepackage{microtype}      % microtypography
\usepackage[dvipsnames,table]{xcolor}         % colors
\usepackage{comment}        % comment out unused sections
\usepackage{amsmath}
\usepackage{graphicx}
\usepackage{placeins}
\usepackage{rotating}
\usepackage{multirow}       % multi row and columns in tables

\title{Training Deep Spiking Auto-encoders without Bursting or Dying Neurons through Regularization}

\author{%
  Justus F.~Hübotter %\thanks{Use footnote for providing further information about author (webpage, alternative address)---\emph{not} for acknowledging funding agencies.} \\
  \\
  Donders Institute for Brain, Cognition and Behaviour\\
  Department of Artificial Intelligence\\
  Radboud University\\
  the Netherlands\\
  \texttt{justus.huebotter@donders.ru.nl} \\
  \AND
  Pablo Lanillos\\
  Donders Institute for Brain, Cognition and Behaviour\\
  Department of Artificial Intelligence\\
  Radboud University\\
  the Netherlands
  \AND
  Jakub M. Tomczak\\
  Department of Computer Science\\
  Vrije Universiteit Amsterdam\\
  the Netherlands
}

\begin{document}

\maketitle

\begin{abstract}

Spiking neural networks are a promising approach towards next-generation models of the brain in computational neuroscience. Moreover, compared to classic artificial neural networks, they could serve as an energy-efficient deployment of AI by enabling fast computation in specialized neuromorphic hardware. However, training deep spiking neural networks, especially in an unsupervised manner, is challenging and the performance of a spiking model is significantly hindered by dead or bursting neurons. Here, we apply end-to-end learning with membrane potential-based backpropagation to a spiking convolutional auto-encoder with multiple trainable layers of leaky integrate-and-fire neurons. We propose bio-inspired regularization methods to control the spike density in latent representations. In the experiments, we show that applying regularization on membrane potential and spiking output successfully avoids both dead and bursting neurons and significantly decreases the reconstruction error of the spiking auto-encoder. Training regularized networks on the MNIST dataset yields image reconstruction quality comparable to non-spiking baseline models (deterministic and variational auto-encoder) and indicates improvement upon earlier approaches. Importantly, we show that, unlike the variational auto-encoder, the spiking latent representations display structure associated with the image class.

\end{abstract}

% =SECTION===============================
\section{Introduction}
% Spiking neural networks (SNNs) may be the next-generation of artificial neural networks (ANNs) and are increasingly becoming
Spiking neural networks (SNNs) are biology-inspired artificial neural networks (ANNs) that have become important tools for both, artificial intelligence (AI) and computational neuroscience research~\cite{Roy2019}. In contrast to other types of ANNs, SNNs are promising candidates for scalable deployment of AI due to their low-energy, low-latency, event-driven computations in specialized neuromorphic hardware \cite{Schuman2017, pfeiffer2018deep, Roy2019}. The internal dynamics of spiking neurons naturally include the notion of time and therefore SNNs combine key benefits from connectionism and dynamicism approaches. %which itself is given increasing attention in the maturation process of neuroscience as a whole~\cite{grondin2010timing}.

% What are they used for
%Spiking neuron models, such as leaky integrate-and-fire (LIF) neurons \cite{Herz2006}, integrate input over time and emit temporally sparse spike events. Therefore, they constitute more detailed models of biological neurons compared to the ones commonly applied in ANNs. SNNs equipped with LIF neurons enable the investigation of learning dynamics based on spike rates and precise spike timing in the brain, as well as deployment of deep learning in specialized low-latency, low-energy neuromorphic hardware \cite{Schuman2017}. Generally, SNNs are applicable to the same tasks as ANNs with arbitrary (hierarchical) network architecture, but achieving best performance when temporal coding is relevant~\cite{davies2021advancing}. Similar to early ANN research, the focus is currently often on shallow feed-forward networks to solve computational benchmark tasks, such as MNIST classification \cite[for a recent comparison please see][]{Illing2019}.
%In contrast to ANNs, SNNs are promising candidates for scalable deployment of AI due to their low-energy, low-latency, event-driven computations in specialized neuromorphic hardware \cite{Schuman2017,pfeiffer2018deep,Roy2019}. 

Spiking neuron models, such as leaky integrate-and-fire (LIF) neurons \cite{Herz2006}, integrate input over time and emit temporally sparse spike events. Therefore, they constitute more detailed models of biological neurons compared to standard ANNs. Generally, SNNs are applicable to the same tasks as ANNs with arbitrary (hierarchical) network architecture, such as representation learning, optimization, or closed-loop systems, and they improve the performance several orders of magnitude when temporal coding is relevant compared to non-spiking networks~\cite{davies2021advancing}. %SNNs enable the deployment of deep learning in specialized low-latency, low-energy neuromorphic hardware~\cite{Schuman2017}, as well as the investigation of learning dynamics based on spike rates and precise spike timing in the brain. 
Despite the new advances in neuromorphic computing~\cite{hunsberger2016training,davies2021advancing}, similarly to early ANN research, SNN studies often focus on shallow feed-forward networks to solve computational benchmark tasks, such as MNIST classification---for a recent comparison see~\cite{Illing2019}. However, deep network structures are desirable to enable hierarchical information abstraction and integration.

% Why is this relevant
The auto-encoder (AE) neural model aims to learn a data-driven representation of the input information in abstracted and compressed form without performing any additional "cognitive" task on this representation (such as classification). Specifically, probabilistic latent representations of variational auto-encoders (VAEs) have recently proven useful in generative processes \cite{han2019variational, razavi2019generating}, (model-based) reinforcement learning \cite{hafner2020mastering} and adaptive control \cite{Meo2021}. While the representation of information in the latent space of non-spiking (deterministic and probabilistic) AE models is well studied~\cite{mathieu2019}, how relevant findings relate to deep spiking models, however, is not always obvious. To the best of our knowledge, such deep spiking AEs (SAEs) supporting unsupervised end-to-end learning has not been properly studied in current literature. Earlier attempts to unsupervised learning in SNNs rely on layer-wise training, require additional image preprocessing, and show poor image reconstruction quality \cite{Burbank2015,Panda2016}.

% Proposing a solution
Here, we aim to address this knowledge gap by applying membrane potential-based backpropagation to a convolutional SAE network with multiple trainable layers of LIF neurons for unsupervised learning. Training deep SNNs is challenging due to their non-differentiable character (i.e., LIFs produce discrete activation events) \cite{Bellec2020,hunsberger2016training,  Lee2019, Neftci2019, OConnor2018, OConnor2016}. %A common issue in SNNs is that some neurons remain active most of the time or, on the contrary, inactive \cite{blalock2020state, Lee2016, qiao2019neural}. As a result, both behaviors are undesirable and hinder final performance of a model. 
Importantly, the threshold-and-reset mechanism of LIF neurons makes SNNs vulnerable to dying neurons (no activation across input stimuli) and bursting neurons (very high firing rates up to constant activation across input stimuli). This is inefficient from an information-theoretic perspective. Both, dead and constantly bursting neurons, cannot contribute actively to a differentiated representation of encoded stimuli and hinder the final performance of a model. Furthermore, a well-known computational bottleneck, in SNN implementations, appears when all neurons fire at the same time. Therefore, controlling the structure of the spiking latent representations is critical. 

The goal of this paper is three-fold: (i) training a deep spiking auto-encoder end-to-end with backpropagation; (ii) avoiding undesirable neuron behavior of constant activity (bursting) and constant inactivity (dead neurons) with appropriate regularization methods; and (iii) comparing spiking model performance and latent representation against non-spiking baseline models. The spiking auto-encoder model was evaluated against two non-spiking baseline model types, namely, a deterministic AE, and a VAE on the MNIST dataset~\cite{lecun1998gradient}, a typically used benchmark in SNN unsupervised learning. While reconstructing digits may be considered a simple task in terms of performance, the results show significant insights for understanding training and representation learning in auto-encoding SNNs. All code is available on github\footnote{\hyperlink{https://github.com/jhuebotter/SpikingVAE}{https://github.com/jhuebotter/SpikingVAE}}.

% =SECTION===============================
\section{Methods}

\subsection{Problem Statement}
\begin{figure}[h!]
    \centering
    \includegraphics[width=0.9\textwidth,height=190px]{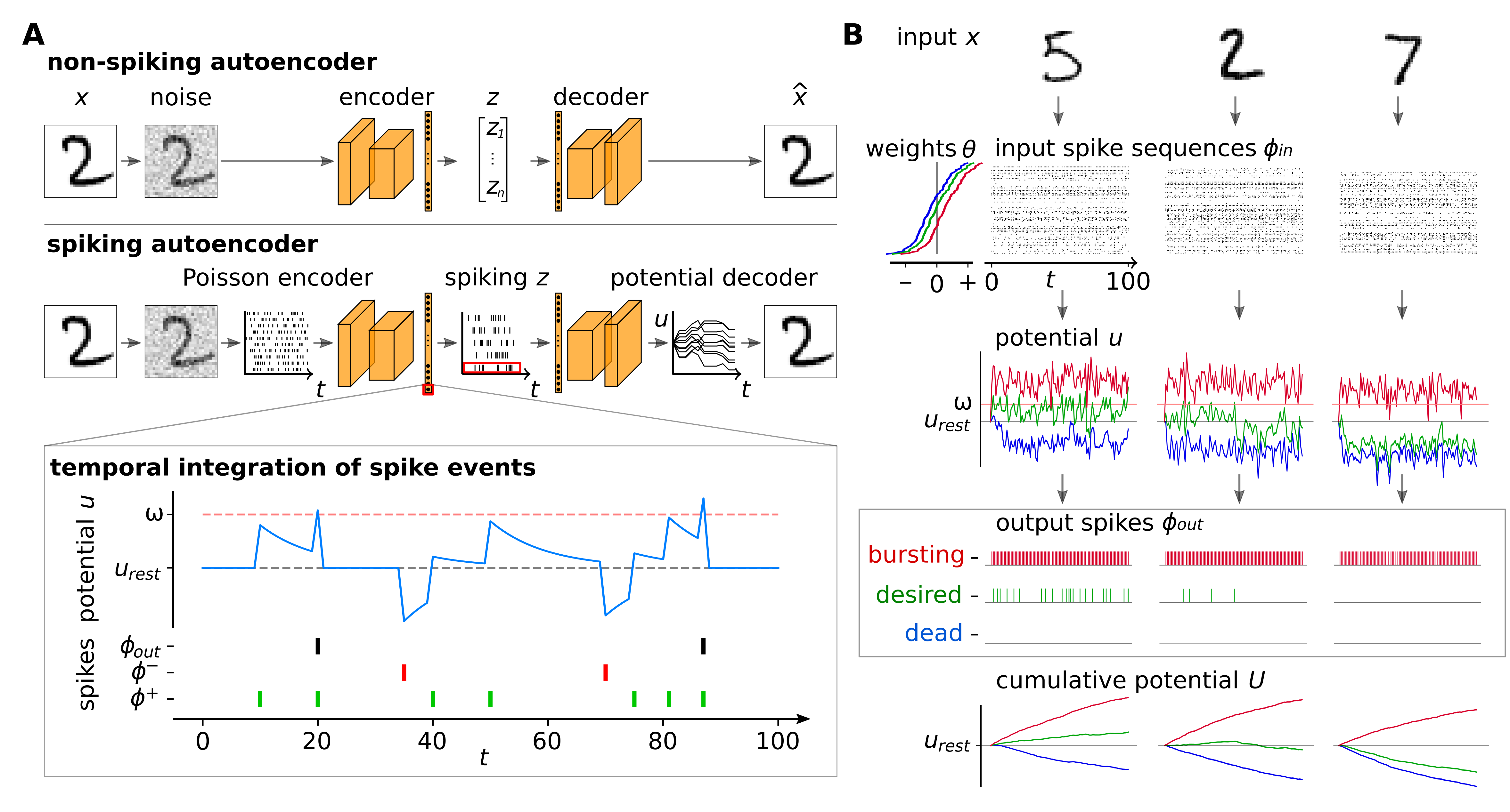}
    \caption{\textbf{Spiking auto-encoder}. \textbf{A} Information processing schematic shows the difference between spiking and non-spiking models. In contrast to non-spiking models, spiking neurons integrate incoming spike events over time in a decaying membrane potential variable $u$. \textbf{B} Desired spiking neuron stimulus-dependent responses (green) and undesired bursting (red) or dead (blue) behaviors. Three spiking neurons, with varying weights $\theta$ (slightly shifted in the mean), are stimulated by three different digits input $\phi_{in}$. While the green neuron presents input-dependent membrane potential $u$ and sparse spiking output $\phi_{out}$, the red neuron displays very high firing rates across inputs and the blue neuron shows no activation across inputs. As the training uses the maximum of the cumulative membrane potential $U$ to approximate $\phi_{out}$ during loss calculation, it remains differentiable with respect to the network weights $\theta$. We propose regularization on $U$ and $\phi$ to balance neural responses and to obtain the desired response (green).
    %Three spiking neurons receive the same input sequences $\phi_{in}$, which are multiplied by the weights $\theta$ which are slightly shifted in the mean. The respective potential $u$ and spiking output $\phi_{out}$ are therefore different, resulting in constant bursting (red), desired stimulus-dependent responses (green), or constant inactivity (blue). The maximum of the cumulative membrane potential $U$ is used to approximate $\phi_{out}$ during loss calculation, as it remains differentiable with respect to the network weights $\theta$. We use regularization on $U$ and $\phi$ to balance neural responses similar to the green curve.
    }
    \label{fig:model_architecture}
\end{figure}

We denote the input by $x$ and a latent representation by $z$. In the AE setup, $x$ is mapped to $z$ by an encoder network, and a reconstruction $\hat{x}$ from $z$ is computed by a decoder network. The encoder can be described as a function $e$ with a set of parameters $\theta_e$, $z = e(x; \theta_e)$, and the decoder network is described by function $d$ with parameters $\theta_d$, $\hat{x} = d(z; \theta_d)$. The auto-encoder can therefore be summarized as $\hat{x} = d(e(x; \theta_e); \theta_d) = f(x; \theta)$. The aim of the learning process is to find values of parameters that minimize the reconstruction loss $\theta^* = \arg\min_{\theta}\mathcal{L}(x, f(x; \theta))$, where $\mathcal{L}_{rec} \equiv \mathcal{L}(x, \hat{x})$ is, e.g., the $\ell_2$ distance averaged over $N$ datapoints:
\begin{equation}
    \label{eq:reconstruction_loss}
    \mathcal{L}_{rec} = \dfrac{1}{N} \sum_{n=1}^N (x_n - \hat{x}_n)^2 .
\end{equation}
%where $N$ is the batch size.
%$\{\theta_e^*, \theta_d^*\} = \arg\min_{\theta_e, \theta_d}\mathcal{L}(x, d(e(x; \theta_e); \theta_d))$.
In this work, we propose to utilize spiking neural networks in the AE architecture. As a result, unlike standard AEs and VAEs, we will use binary latent variables, i.e., spikes, that will be represented by a matrix with one dimension being temporal. Since we use SNNs, we need to modify the input to the encoder (image-to-spike), and the output of the decoder (potential-to-image). The proposed architecture is schematically presented in \autoref{fig:model_architecture}A.

Controlling the spike activity patterns in SNNs, besides being important for unsupervised learning, is major to achieve the best performance when deploying in neuromorphic hardware. On the one hand, sparse representations with respect to its effects and underlying mechanisms is a desirable property~\cite{Ahmad2019, Chen2017, Cramer2020, gale2019state, Goh2010, Levy2004, QuianQuiroga2010}. On the other hand, if some neurons are never or constantly active, the model capacity drops significantly. \autoref{fig:model_architecture}B describes two common issues that cripple training and decrease the final performance of SNNs: dying and bursting neurons. 
\vspace{-0.3cm}
\paragraph{Dying neurons}
The dying neuron problem refers to neurons that are consistently silent or inactive independent of the input. This is also a common issue in ANNs with ReLU activation function~\cite{lu2019dying}. During weight initialization or update, it regularly happens that certain neurons no longer receive sufficient input to elicit output from the activation function. In this case, and in the absence of spontaneous neuronal activity, the gradients for any following error backpropagation calculation for such neurons are $0$, and no learning occurs. This essentially renders the neuron permanently inactive, or "dead"---See \autoref{fig:model_architecture}B blue neuron. In some cases, this is desired, and dead neurons can be pruned to reduce network size during or after learning \cite{blalock2020state}. Alternatively, various methods have been proposed to prevent the death of too many or even any neurons, e.g. by adjusting input weights or firing thresholds dynamically \cite{Lee2016} or neural rejuvenation \cite{qiao2019neural}.

\vspace{-0.2cm}
\paragraph{Bursting neurons} In contrast to dead neurons, the bursting neuron problem is given by the alternative extreme behavior of spiking neurons through constant activity---See \autoref{fig:model_architecture}B red neuron. Although some input is required to elicit a response in LIF neurons due to the lack of a bias term, bursting behavior may occur across inputs and therefore independent of the stimulus type. 

Arguably, both dead and bursting neurons no longer participate in the representation, encoding, or decoding of information in the network, and when this occurs at a large scale, it can significantly change the network's information storage and processing capacity. To avoid dying and bursting neurons, we propose (loosely) biologically inspired regularization methods as additional loss terms in order to leverage the already applied gradient-based weight update mechanism. The considered additional error terms are posing a soft constraint on network behavior rather than a hard limit and are described in more detail in \autoref{sec:regularization}. Before that, we outline the main components of the SAE, such as a LIF neuron model, the image-to-spike encoder and the potential-to-image decoder.

\subsection{Spiking auto-encoder}

\paragraph{Leaky Integrate and Fire Model} The most influential and widely used model of neuron dynamics is the LIF model \cite{Herz2006, Paulin2014, Zambrano2019}. It integrates information over time in a leaky membrane-potential-inspired variable and produces binary spike events as output via a non-linear threshold-and-reset mechanism. The membrane potential dynamics and nonlinear activation function of the LIF neuron model are described by a set of equations solved at discrete time steps $t \in [1,\,T]$. This is more computationally expensive than an event-based computation, but allows for the modeling of the sub-threshold dynamics of each neuron's membrane potential as opposed to only their time of spiking. The membrane potential $u(t)$ of any LIF neuron $i$ is recursively dependent on itself (after potential reset) as well as incoming spike events from all projecting neurons $j$ weighted by synaptic strength parameters $\theta_{ij}$ as:
\begin{equation}
    \label{eq:membrane_potential}
    u_{i}(t)=\tau u_{i}'(t-1) + \sum_j{\theta_{ij}\phi_j(t)}\,,
\end{equation}
where $\tau \in [0,\,1]$ is a constant hyper-parameter for decay, $\theta_{ij}$ is the weight or synaptic strength of the connection between neuron $i$ and $j$. Note that no bias term is used in this model. $\phi_j(t)$ is the binary output of neuron $j$ and depends on the membrane potential $u_j(t)$ and a threshold parameter $\omega = 1$:
\begin{equation}
    \label{eq:threshold}
    \phi_j(t) =
    \begin{cases}
            1, &         \text{if } u_j(t)\ge \omega,\\
            0, &         \text{otherwise}.
    \end{cases}
\end{equation}
Finally, if a neuron $i$ emits a binary spike event of $\phi_i(t) = 1$, the membrane potential is reset to its initial and resting potential $u_i'(t=0) = u_{rest} = 0$ by:
\begin{equation}
    \label{eq:reset}
    u_i'(t) =
    \begin{cases}
            u_{rest}, &         \text{if } u_i(t) \ge \omega,\\
            u_i(t), &    \text{if } \omega > u_i(t) \ge -\omega,\\
            -\omega, &        \text{if } u_i(t) < -\omega.
    \end{cases}
\end{equation}
The lower bound is set on the membrane potential $u_i(t)$ to avoid strongly negative values, which cause undesired network behavior. %The details of the encoding of static input images into a temporal sequence as well as the reversed decoding process are described in the following sections.

\paragraph{Information encoding} Before passing any input image $x$ through a model, the image is corrupted by adding uniformly distributed noise to each pixel according to:
\begin{equation}
    \label{eq:noise}
    x'_{ij} = (1 - \epsilon) x_{ij} + \epsilon \xi\,,
\end{equation}
with $\epsilon \in [0,\,1]$ being a hyper-parameter and $\xi \sim \mathcal{U}(0,\,1)$.

%There are a number of ways to encode a static input into a temporal sequence. The three approaches predominantly used in literature are: current encoding, Poisson encoding, and time-to-first-spike encoding. The first two options are typically interpreted as rate codes, the latter encodes information only in time. All methods allow the addition of a second source of noise which varies the signal over time if desired, but this step is omitted here for simplicity.

To encode the static input into a temporal sequence, we use Poisson encoding\footnote{Poisson encoding is one of the most common approaches along with current and time-to-first-spike encoding.} \cite{Burbank2015, Panda2016} where the static input is transformed into a series of binary events for each input feature, where the density of spikes (1s) is approximately proportional to the magnitude of the respective input feature. This can be simulated through a stochastic process where:
\begin{equation}
    \label{eq:poisson_encoding}
    x''_{ij}(t) = 
    \begin{cases}
            1, &         \text{if } sx'_{ij} > r(t),\\
            0, &         \text{otherwise}.
    \end{cases}
\end{equation}
with $r(t) \sim \mathcal{U}(0,\,1)$ for every discrete time step $t \in [1, T]$ and $s$ being a constant scaling factor.% hyper-parameter.

\paragraph{Information decoding} The outputs of the SAE model are temporal sequences of spike events $\phi_i(t)$ as well as cumulative leaky membrane potentials $U_i(t)$, as an approximation for the respective neurons firing rate. %\footnote{Conversely, both baseline model types AE and VAE directly output $\hat{x}$ in the same shape as the original $x$ with each feature due to the activation function of the respective output layer.}. 
The latter is defined as:
\begin{equation}
    \label{eq:cum_mp}
    U_i(t) = \tau U_i(t-1) + u_i(t)\,.
\end{equation}
To obtain the reconstructed $\hat{x}$, we utilize the maximum value of $U_i(t)$:
\begin{equation}
    \label{eq:max_decoder}
    \hat{x} = \dfrac{1}{s}\max_{t \in [t_{min},\,T]} U_i(t)\,,
\end{equation}
with $s$ being a constant scaling factor hyper-parameter. This bounds $\hat{x}_{ij} \in [0,\,\infty)$ just like the ReLU activation function and does not demand extended network activity to generate valid output. Moreover, it allows to give the network an initialization period by setting $t_{min} > 0$, For this work, we set that $t_{min} = 0$. This cumulative membrane potential can be used for training. However, using the maximum $\max U$ limits precise time-of-spike coding \cite{Zhang2019}.

%Unfortunately, a decoding method which computes $\hat{x}$ based on time-to-first-spike of the output layer in a way that remains differentiable with respect to the network parameters $\theta$ while computing discrete time steps is challenging.

\paragraph{Training with membrane-potential backpropagation} The threshold or step activation function has a gradient of 0 with respect to the weights, thus, it does not allow for backpropagation. Instead, the decaying cumulative membrane potential $U(t)$ of the output layer is used as a proxy for neural activity. During backpropagation, the threshold-and-reset mechanism is considered noise on the potential variable $u(t)$ and is ignored by a straight-through estimate \cite{Bengio2013} as proposed for supervised learning implementations of spike-based backpropagation \cite{Lee2019,Zenke2018}.

\subsection{Regularization methods}
\label{sec:regularization}

Here, we avoid dying and bursting neurons using soft constraints via regularization \cite{Cramer2020, Goh2010}. The proposed regularizers for the SAE are conceptually based on different metabolic costs of processes within biological neurons to maintain the membrane potential or generate action potentials \cite{Attwell2001}. These can be categorized as regularization of weights, neuron potential, and neuron activity.

\paragraph{Weight regularization} We penalize for synaptic connections of nonzero strength. This method represents a metabolic upkeep cost for each synaptic connection:
\begin{equation}
    \label{eq:l2}
    \textcolor{ForestGreen}{\mathcal{L}_{L2}} = \sum_{l}^{L}\sum_{i}^{I}\sum_{j}^{J}\theta_{lij}^2\,.
\end{equation}
where all model parameters $\theta$ are squared and summed for all layers $l \in [1, L]$. The respective gradient is directly dependent on the value of the synaptic strength. This regularization method is commonly applied in statistical models and machine learning and yields generally smaller parameter values with a focus on penalizing specifically largely positive or negative values. It is conceivable that in biological neurons this mechanism has a physiological counterpart, in the sense that the presence of a synapse comes with metabolic upkeep cost, which increases depending on the strength of the synaptic connection. This concept of a metabolic cost extends to neuron behavior and is the basis for the following regularization terms. 

\paragraph{Potential regularization} We penalize for membrane potential deviating from the resting state, and is therefore specific to SNN models. In this context, however, it fulfills an immensely important function, as it prevents neurons getting trapped in a state of constant inhibition. In SNNs the problem of dying neurons can be avoided by directly penalizing the membrane potential deviation from resting state, independent of the respective neuron's spiking output. Mathematically, this is defined as the two terms:
\begin{equation}
    \label{eq:p1}
    \textcolor{blue}{\mathcal{L}_{P1}} = \sum_{l}^{L}\sum_{i}^{I}|U_{li}(t_{max})|\quad  \text{and} \quad 
    \textcolor{blue}{\mathcal{L}_{P2}} = \sum_{l}^{L}\sum_{i}^{I}U_{li}(t_{max})^2\,.
\end{equation}
The underlying assumption is that there is a mechanism available, which allows a biological neuron to sense and keep track of recent membrane potential deviation trends and to adjust its sensitivity to excitation and inhibition to maintain a balance between these two driving forces. This mechanism does not seem far-fetched in the light of the multitude of dynamic processes which regulate short and long-term plasticity in neurons both locally at any synapse as well as globally via epigenetic mechanisms.

\paragraph{Activity regularization} We penalize for nonzero neuron output. In this case, each neurons spike output is considered costly, expressed by the term:
\begin{equation}
    \label{eq:a1}
    \textcolor{red}{\mathcal{L}_{A1}} = \sum_{l}^{L}\sum_{i}^{I}\dfrac{1}{T}\sum_{t=1}^{T}\phi_{li}(t)\,,
\end{equation}
Applying this regularization term reduces the number of spike events present in the network activity and was inspired by previous works \cite{Cramer2020}. However, as this activity can only be of positive value, this method by itself has the potential to reduce activity to the point of inducing the dead neuron problem. Therefore, the combination of this approach with potential regularization is expected to both avoid dead neurons and induce temporally sparse activity patterns.

In our AE architecture, this constraint is useful to be applied to a single layer. For applying the activity regularization to the most compressed latent spikes representation (e.g., layer $l = 3$) we define $\mathcal{L}_{A1_{l=3}}$.

\paragraph{Regularized SAE general objective function} Using all regularization terms the SAE takes the following training objective.
\begin{equation}
    \label{eq:loss_SAE}
    \mathcal{L}_{SAE} = \mathcal{L}_{rec} + \textcolor{ForestGreen}{l_2\mathcal{L}_{L2}} + \textcolor{blue}{p_1\mathcal{L}_{P1}} + \textcolor{blue}{p_2\mathcal{L}_{P2}} + \textcolor{red}{a_1\mathcal{L}_{A1}} %+ %n_1\mathcal{L}_{N1}\,,
\end{equation}
Where each regularization term has an associated weight hyper-parameter. Depending on the regularization terms used, the SAE model will present different activation behaviors---See \autoref{tab:compared_models}. A SAE trained with $\mathcal{L}_{P1}$, $\mathcal{L}_{P2}$ and $\mathcal{L}_{A1}$ will force sparsity in the latent representation (hereafter referred to as \textit{SAE-sparse}). A SAE trained with $\mathcal{L}_{P2}$ and $\mathcal{L}_{A1_{l=3}}$ will provide dense activation without bursting or dying neurons (hereafter referred to as \textit{SAE-dense}).

For consistency, we define the loss for the baselines using the same notation. The deterministic AE's training objective is: $\mathcal{L}_{AE} = \mathcal{L}_{rec} + \textcolor{ForestGreen}{l_2\mathcal{L}_{L2}}$.
The VAE's loss function takes the following form: $\mathcal{L}_{VAE} = \mathcal{L}_{rec} + \beta\mathcal{L}_{KL} + \textcolor{ForestGreen}{l_2\mathcal{L}_{L2}}$, where $\mathcal{L}_{KL}$ is the Kullback-Leibler divergence term \cite{kingma2013auto}.

% =SECTION===============================
\section{Experiments}
We compared three versions of the proposed SAE with a deterministic AE and a VAE---Summarized in \autoref{tab:compared_models}. Both baseline models use the rectified linear unit (ReLU) activation function in all layers, with the exception of the layer constructing the latent representation in the VAE. In contrast to the baseline, all SAE models use a LIF mechanic as a nonlinear activation function. Unlike SAE, the baselines have a learned bias parameter for each neuron. All models were trained on the MNIST dataset for 10 epochs with 5 repetitions. Training and architecture implementation details, hyperparameters analysis and further studies, such as bottleneck impact and latent representation clustering, can be found in the Appendix.

\begin{comment}
\begin{table}[hbpt!]
    \centering
    \caption[Regularization weights]{Overview of compared models and their respective regularization term weights.}
    \begin{tabular}{l l}
        \textbf{Model name} & \textbf{Non-zero regularization weights} \\
        \hline
        SAE & no regularization\\
        SAE-sparse & $p_1 = 0.005$; $p_2 = 0.005$; $a_1 = 0.01$ \\
        SAE-dense & $p_2 = 0.01$; $a_{1,l=3} = 0.01$\\
        AE & no regularization\\
        AE$_{l2}$ & $l_2 = 0.00001$\\
        VAE & $l_2 = 0.01$; $\beta = 1.0$ \\
        $\beta$VAE & $l_2 = 0.01$; $\beta = 0.1$
    \end{tabular}
    \label{tab:compared_models}
\end{table}
\end{comment}

\begin{table}[hbtp!]
    \centering
    \caption[Regularization weights]{Overview of compared models and their respective regularization term weights as well as the reconstruction error and ratio of average intra-class to inter-class latent representation distances.}
    \begin{tabular}{l l r r r r}
        \toprule
        \textbf{Model name} & \textbf{Non-zero regularization weights} & \multicolumn{2}{l}{\textbf{$\mathcal{L}_{rec}$} \small(mean $\pm$ std)}\normalsize & \multicolumn{2}{l}{$\overline{s}_{intra} / \overline{s}_{inter}$}\\
        \midrule
        SAE         &  no regularization                                             & 38.09 & $\pm$ 8.98 & 0.65 & $\pm$ 0.07 \\
        SAE-sparse  & $p_1 = 0.005$; $p_2 = 0.005$; $a_1 = 0.01$    & 11.58 & $\pm$ 0.89 & 0.74 & $\pm$ 0.01 \\
        SAE-dense   & $p_2 = 0.01$; $a_{1_{l=3}} = 0.1$                     & 6.75  & $\pm$ 0.26 & 0.78 & $\pm$ 0.00 \\
        AE          & no regularization                                     & 7.89  & $\pm$ 4.83 & 0.65 & $\pm$ 0.08 \\
        AE$_{l2}$   & $l_2 = 0.00001$                               & 4.67  & $\pm$ 0.98 & 0.69 & $\pm$ 0.04 \\
        VAE         & $l_2 = 0.01$; $\beta = 1.0$                   & 19.09 & $\pm$ 0.66 & 0.97 & $\pm$ 0.00 \\
        $\beta$VAE  & $l_2 = 0.01$; $\beta = 0.1$                   & 5.94  & $\pm$ 0.21 & 0.94 & $\pm$ 0.00 \\
        \bottomrule
    \end{tabular}
    \label{tab:compared_models}
\end{table}

All models used in this work use the same architecture depicted in \autoref{fig:model_architecture}A. It consists of 6 layers in total, 3 for the encoder and 3 for the decoder. The input is first passed through two convolutional layers for feature map generation with 16 and 32 5x5 kernels respectively \cite{krizhevsky2012imagenet}. %These convolutional kernels can each be understood as a learned receptive field which is applied to all positions of the input space to generate a feature map . 
The final layer of the encoder takes in a flattened representation of these feature maps and is fully connected, resulting in a flat latent representation of the data of size $n_z$. In the decoding process, this latent representation is passed through another fully connected layer and two deconvolutional layers which mirror the encoder so the output size matches that of the input. While SAE does not use regularization terms, SAE-sparse uses the potential and the activation regularization in all layers. Finally, the SAE-dense only uses the squared potential penalty term and the activity regularization on layer 3 (deepest layer).

Table~\ref{tab:compared_models} summarizes the compared models and their reconstruction performance (lower is better) as well as the intra-inter class distance ratio (lower is better), as a measure of clustering. The best performance was reported for the $\text{AE}_{l2}$ and SAE-dense achieved the best spiking performance. This indicates the strong impact of regularization in SAEs unsupervised learning, similarly to VAEs. However, unlike VAEs, SAEs presented better class representation clustering---See Appendix \ref{sec:appendix}.5.

\subsection{Qualitative Model Comparison}
\label{sec:qualitative_results}

\begin{figure}[hbtp!]
    \centering
    \includegraphics[width=0.9\textwidth, height=240px]{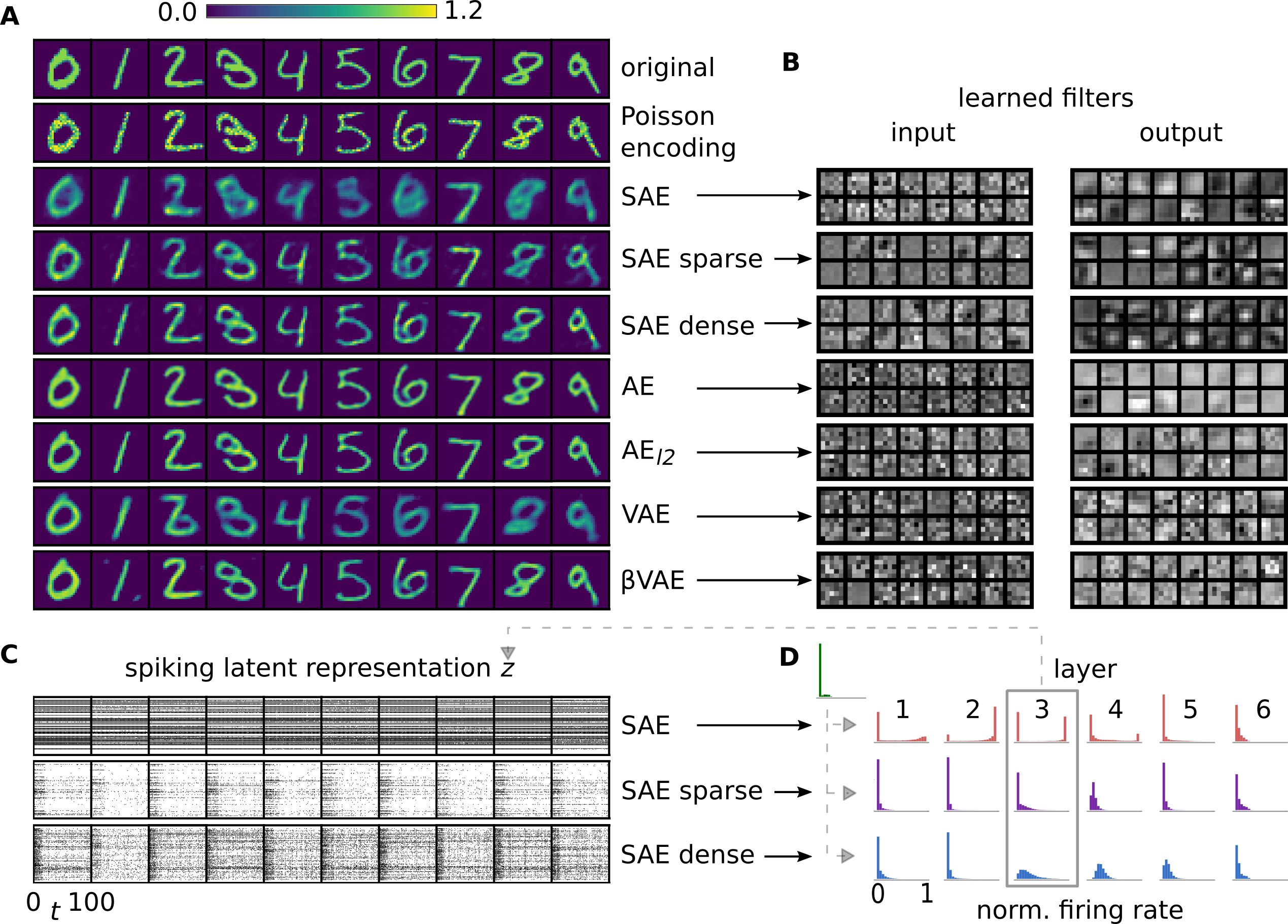}
    \caption{\textbf{Qualitative model comparison}. \textbf{A} Input, Poisson encoding, and model reconstruction of MNIST images for all models. \textbf{B} Comparison of the models learned receptive fields for input and output layers. \textbf{C} Comparison of latent representation of the MNIST examples from panel A across spiking models. \textbf{D} Normalized firing rate distributions across layers of spiking models and the Poisson encoding (top-left in green). Examples of all panels were generated from the model with the respective lowest reconstruction error across the 5 repetitions.}
    \label{fig:qualitative_results}
\end{figure}

% example image reconstructions
We evaluated the reconstruction quality, the learned filters and the SAE latent representation. \autoref{fig:qualitative_results}A shows 10 example MNIST images, one per class, as well as their Poisson encoding and reconstruction by the spiking models and the regularized baseline models. The Poisson encoding (summed and normalized over 100 time steps) closely resembles the original images. The unregularized SAE shows poor reconstruction quality which is visibly improved by regularization methods. The SAE-dense reconstruction quality is comparable to the non-spiking baseline models. The example image for digit "3" constitutes an edge-case as slight differences between the original image and reconstruction of the dense SAE and $\beta$VAE result in output that closely resembles an "8". In any case, these results are a clear improvement upon earlier approaches to unsupervised learning with SAEs \cite{Burbank2015, Panda2016} (see also Appendix \ref{sec:appendix}.6).

% learned input and output filters
Each models' 16 convolutional kernels of the input and output layer respectively are expected to learn receptive fields close to Gabor filters to detect lines and edges of different orientations across the MNIST images. \autoref{fig:qualitative_results}B shows that all model types still show little structure in their input filters after 10 training epochs after visual inspection. The output layer filters, however, seem to have learned more structure of the underlying data compared to the input filters. This is especially true for the regularized SAE models with clearly identifiable receptive fields. A possible explanation for the difference between input and output filter clarity may be the stronger learning signal at the end of the network architecture due to vanishing gradients. Conversely, regularized non-spiking baseline models presented poorly recognizable filter structures despite their high image reconstruction quality.

%Potentially the 10 epochs of training were not sufficient for more defined filters to arise in early layers. The fact that the regularized non-spiking baseline models show the highest image reconstruction quality despite poorly recognizable filter structure is inherently unexpected.

% spiking latent representations
\autoref{fig:qualitative_results}C shows the spiking latent representation (layer 3) of the images. The unregularized SAE is a perfect example of bursting and dying neurons. It consistently shows strong bursting behavior in some neurons and inactivity in others across all examples. However, the representation of the SAE-sparse shows a great reduction of spike events per example compared to the unregularized model with no bursting behavior, but still some neurons are consistently inactive. The SAE-dense appears to have a spike density between the unregularized and sparse model with a more balanced distribution of activity across neurons. Here, no dying or bursting neurons are recognizable upon visual inspection. Interestingly, both regularized models show a higher density in spike events in the first 10-20 time steps, after which the firing rate plateaus on a lower level across neurons. This behavior may be that the output image is decoding as the maximum of the cumulative membrane potential of output neurons according to \autoref{eq:max_decoder}. Visual inspection of the respective data for $U(t)$ shows that indeed this maximum is reached reliably within this early time window (data not shown). This finding suggests that fewer time steps may be sufficient to solve the image reconstruction tasks by the SAE models, which is evaluated below in \autoref{fig:comparison}B \& D.

% firing rate histograms for spiking models
The distributions of normalized firing rates across neurons for each of the three spiking models and their layers show an initial increase (layer 1 to 4) and subsequent decrease (layer 5 and 6) of average neuronal activity (\autoref{fig:qualitative_results}D). While the firing rate distributions in the earlier layers of the unregularized SAE show a bi-modal distribution with two clear peaks at the extremes (one for dead and one for bursting neurons) the regularized models show Poisson-like distributions across all layers. The mean firing rates of the dense model are slightly higher compared to the sparse model. Furthermore, the firing rate distribution of layer 6 (output) is well comparable across models, although not as sparse as the Poisson encoded input, suggesting excess spikes in the network output.

% default case: no noise, 100 neurons in latent space, 100 time steps, and decay of 0.99

\subsection{Quantitative Model Comparison}
\label{sec:quantitative_results}

% We evaluated the data efficacy, the batch processing time and the performance, under different variations of the architecture (bottleneck size, input noise) across models (\autoref{sec:data_efficacy}). All quantified results were obtained on the validation set and all reported statistical tests are based on Welch's t-test unless otherwise specified.

%Reconstruction loss and other model behavior descriptors are summarized in \autoref{sec:statistical_results} for the default parameters and in \autoref{sec:information_bottlenecks} under different information bottleneck conditions. All quantified results were obtained on the validation set and all reported statistical test results are based on Welch's t-test unless otherwise specified.

%Some expectations here would be nice?

\begin{comment}
\begin{itemize}
    \item Regularization solves dying and bursting neuron problem
    \item Regularization improves reconstruction
    \item Regularization does not induces sparse distributed representation in space
    \item Regularization induces sparse representation in time (SAE only)
    \item SAE needs less training examples compared to baseline
    \item SAE needs more runtime compared to baseline
\end{itemize}
\end{comment}

\paragraph{Regularized SAE Models Successfully Learn MNIST Reconstruction Task}
\label{sec:statistical_results}
% quantitative analysis

\begin{figure}[h!]
    \centering
    \includegraphics[ height=230px]{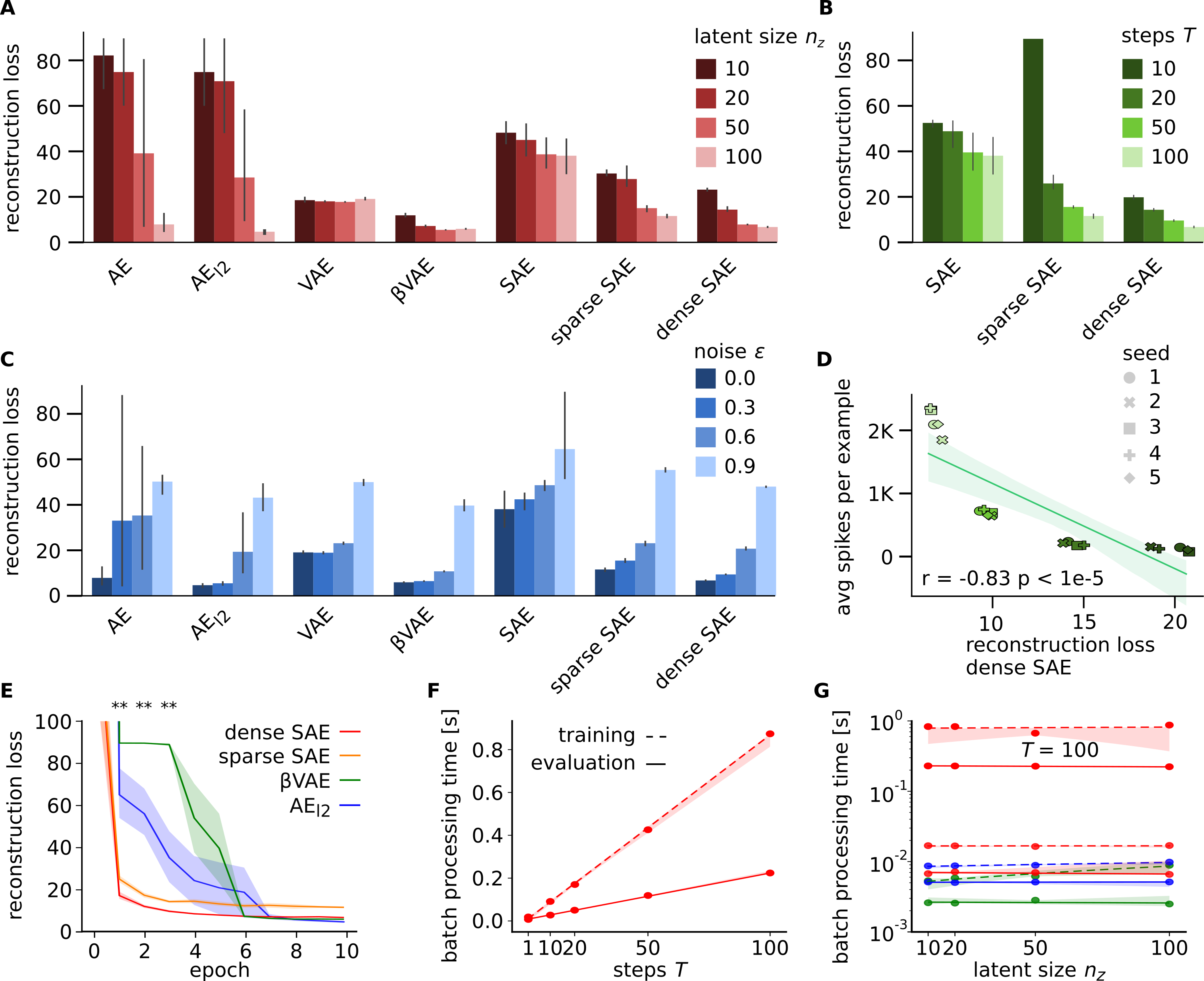}
    \caption[Quantitative model comparison]{\textbf{Quantitative model comparison under information bottleneck conditions.} %The models' image reconstruction error under information bottleneck conditions. 
    Reported data was obtained on the validation set after 10 epochs of training with 5 randomly initialized repetitions (seed). \textbf{A} Reducing the size of the latent representation generally increases model error, except for VAE. Both AE models fail to reliably solve the reconstruction task with $n_z\leq50$. \textbf{B} Decreasing the number of simulated time steps in SAE models generally increases reconstruction error. In contrast to the dense SAE, the sparse model cannot solve the task with $T=10$. \textbf{C} Increasing the noise in the input signal generally increases model error with no tested model capable of solving the reconstruction task with $\epsilon=0.9$. \textbf{D} Dense SAE models reliably show a trade-off between number of avg. spikes per example and reconstruction error when simulated step number is varied. \textbf{E} The regularized SAE models reliably learn the task with fewer training examples compared to baseline. \textbf{F} and \textbf{G} SAE batch processing time scales with the number of simulated time steps but not model size.}
    \label{fig:comparison}
\end{figure}

Regularized SAE models successfully learn the MNIST reconstruction task and achieve comparable performance to the baseline models, even under different information bottleneck conditions such as noise and reduced latent size -- see \autoref{fig:comparison}A \& C. In particular, the SAE dense performs best among the spiking models. Training the \textit{unregularized SAE} for MNIST reconstruction for 10 epochs results in high reconstruction error on the validation set (MSE = 38.09 $\pm$ 8.98), a large proportion of dead layer three neurons (active neuron ratio, ANR = 0.17 $\pm$ 0.20) and strong bursting (constant firing) in other neurons (avg. firing rate of active neurons, AFR = 0.75 $\pm$ 0.11). As predicted, the \textit{SAE-sparse} successfully overcomes the dying neuron problem (ANR = 0.93 $\pm$ 0.04; $p<0.01$) as well as significantly reduces neuronal bursting (AFR = 0.06 $\pm$ 0.02), while improving the reconstruction error (MSE = 11.58 $\pm$ 0.89; $p<0.01$). The ratio of average number of active neurons per example over all active neurons (RAE) (RAE = 0.61 $\pm$ 0.02) suggests a significant increase in spatial sparseness of the latent representation compared to the unregularized model (RAE = 0.95 $\pm$ 0.04; $p<0.0001$). Visual inspection of layer three neurons' average firing rate for 100 examples (see Appendix \autoref{fig:example_clustering} top row center), however, shows that these descriptions are partially deceiving. In fact, a large number of neurons in these models rarely ever emit spike events across inputs, which lets these neurons appear not dead, but their contribution to information representation is questionable. 
% regularization solves dead neuron problem
The \textit{SAE-dense} also shows no dead neurons (ANR = $1.0 \pm 0.0$) but, in contrast to the SAE-sparse, a much more balanced activity across active neurons (see Appendix \autoref{fig:example_clustering} top row right). While SAE-dense shows no spatial sparsity (RAE = 1.0 $\pm$ 0.0), the average firing rate of active neurons (AFR = $0.21 \pm 0.02$; $p<0.001$) is still significantly reduced compared to the unregularized model and image reconstruction error (MSE = $6.75 \pm 0.26$) is significantly lower compared to the other two SAE models ($p<0.001$). 

The unregularized deterministic AE suffers from the dying neuron problem in the latent representation (ANR = $0.22 \pm 0.09$) which is comparable to that of the SAE, but with much lower reconstruction error (MSE = $7.89 \pm 4.83$). AE$_{l2}$, with The weight decay regularization reduces both dead neurons (ANR = $0.41 \pm 0.13$; $p<0.05$) and reconstruction error (MSE = 4.67 $\pm$ 0.98) compared to the unregularized AE. While the probabilistic VAE shows no dead neurons in the latent representation due to the lack of the ReLU activation function, the reduction of the KL-divergence loss term weight $\beta$ from 1.0 to 0.1 significantly improves reconstruction quality (VAE MSE = $19.09 \pm 0.66$; $\beta$VAE MSE = $5.94 \pm 0.21$; $p<0.0001$). Although the image reconstruction quality of all regularized models is comparable (\autoref{fig:qualitative_results}A) the reconstruction error of the AE$_{l2}$ is still lower than $\beta$VAE. 

\paragraph{Data Efficacy and Batch Processing Time}
\label{sec:data_efficacy}

% data efficacy and training time
The regularized spiking models need fewer training examples to produce recognizable output images compared to baseline (see \autoref{fig:comparison}E). While both sparse and dense SAE show good reconstruction quality reliably after a single training epoch, AE$_{l2}$ and $\beta$VAE take between 1 and 7 or 4 and 6 epochs respectively. This group difference in reconstruction error is significant for epochs 1 through 3 ($p<0.0001$ Kruskal–Wallis). With regard to time, however, due to the simulation of and backpropagation through 100 time steps, the spiking models take much longer for processing a single batch compared to baseline (\autoref{fig:comparison}G). The processing time scales linearly with the number of simulated time steps $T$ (\autoref{fig:comparison}F), but not with model size (\autoref{fig:comparison}G). This difference is important because it constitutes the key limitation to the scalability of this approach. Backpropagation through time requires that the spiking output and membrane potential of all neurons in the network are stored in memory for each simulated time step. This linear growth of memory usage effectively prevents the simulation of longer sequences. How learning on a continuous stream of data can be realized with this method is still challenging. Alternative learning algorithms must overcome this limitation, for example by relying on approximate transient signals indicating neuron activity. An interesting candidate for this task may be e-prop \cite{Bellec2020}.

% =SECTION===============================
\section{Conclusion}

The results of this work show that the described SAE models consisting of multiple fully trainable layers of LIF neurons are successfully performing end-to-end learning via error backpropagation. We showed that regularization of the membrane potential and spiking output of LIF neurons is an effective method to considerably improve SAE image reconstruction error while avoiding neural bursting behavior as well as dying neurons. Regularized SAE models successfully learn the MNIST reconstruction task and specifically, the SAE-dense shows reconstruction quality comparable to non-spiking baseline models while requiring less training data but longer processing time. These results constitute an improvement upon earlier approaches to unsupervised learning in SNNs \cite{Burbank2015, Panda2016}. Furthermore, hierarchical clustering of the latent representations reveals similarities between examples of the same stimulus category in regularized SAE models despite the absence of label information during training. This representation structure quality is not present in the probabilistic VAE models, a difference to be kept in mind when using one or the other as a model for early perception. Finally, layer three neuron firing rates are correlated across examples and organized in functional subgroups with no strict borders suggesting a distributed population code with weak redundancies. 

Our proposed regularized spiking-autoencoder allows us to control the structure and density of the spiking latent representations. This study is essential for achieving optimal performance in neuromorphic implementations of unsupervised learning models.
%Based on these results, all three goals listed in the introduction of this work are deemed successfully reached. 

% \paragraph{Broader Impact}\\
% This work provides a novel approach ...
% \cite{roy2019towards}. \cite{pfeiffer2018deep}.

%\paragraph{Acknowledgments} The authors would like to extend their gratitude towards ANONYMIZED for providing valuable feedback in the writing process.

%%%%%%%%%%%%%%%%%%%%%%%%%%%%%%%%%%%%%%%%%%%%%%%%%%%%%%%%%%%%
\bibliographystyle{plain}
\bibliography{literature}

\begin{thebibliography}{10}

\bibitem{Ahmad2019}
Subutai Ahmad and Luiz Scheinkman.
\newblock How can we be so dense? the benefits of using highly sparse
  representations.
\newblock {\em arXiv preprint arXiv:1903.11257}, 2019.

\bibitem{Attwell2001}
David Attwell and Simon~B. Laughlin.
\newblock {An Energy Budget for Signaling in the Grey Matter of the Brain}.
\newblock {\em Journal of Cerebral Blood Flow \& Metabolism},
  21(10):1133--1145, oct 2001.

\bibitem{bal2016medium}
Henri Bal, Dick Epema, Cees de~Laat, Rob van Nieuwpoort, John Romein, Frank
  Seinstra, Cees Snoek, and Harry Wijshoff.
\newblock A medium-scale distributed system for computer science research:
  Infrastructure for the long term.
\newblock {\em Computer}, 49(5):54--63, 2016.

\bibitem{Bellec2020}
Guillaume Bellec, Franz Scherr, Anand Subramoney, Elias Hajek, Darjan Salaj,
  Robert Legenstein, and Wolfgang Maass.
\newblock {A solution to the learning dilemma for recurrent networks of spiking
  neurons}.
\newblock {\em Nature Communications}, 11(1):1--15, 2020.

\bibitem{Bengio2013}
Yoshua Bengio, Nicholas L{\'e}onard, and Aaron Courville.
\newblock Estimating or propagating gradients through stochastic neurons for
  conditional computation.
\newblock {\em arXiv preprint arXiv:1308.3432}, 2013.

\bibitem{wandb}
Lukas Biewald.
\newblock Experiment tracking with weights and biases, 2020.
\newblock Software available from wandb.com.

\bibitem{blalock2020state}
Davis Blalock, Jose Javier~Gonzalez Ortiz, Jonathan Frankle, and John Guttag.
\newblock What is the state of neural network pruning?
\newblock In {\em Machine Learning and Systems}, 2020.

\bibitem{Burbank2015}
Kendra~S. Burbank.
\newblock {Mirrored STDP Implements Autoencoder Learning in a Network of
  Spiking Neurons}.
\newblock {\em PLoS Computational Biology}, 11(12):e1004566, 2015.

\bibitem{Chen2017}
Yanqing Chen.
\newblock {Mechanisms of Winner-Take-All and Group Selection in Neuronal
  Spiking Networks}.
\newblock {\em Frontiers in Computational Neuroscience}, 11:20, apr 2017.

\bibitem{Cramer2020}
Benjamin Cramer, Sebastian Billaudelle, Simeon Kanya, Aron Leibfried, Andreas
  Gr{\"u}bl, Vitali Karasenko, Christian Pehle, Korbinian Schreiber, Yannik
  Stradmann, Johannes Weis, et~al.
\newblock Training spiking multi-layer networks with surrogate gradients on an
  analog neuromorphic substrate.
\newblock {\em arXiv preprint arXiv:2006.07239}, 2020.

\bibitem{davies2021advancing}
Mike Davies, Andreas Wild, Garrick Orchard, Yulia Sandamirskaya, Gabriel
  A~Fonseca Guerra, Prasad Joshi, Philipp Plank, and Sumedh~R Risbud.
\newblock Advancing neuromorphic computing with loihi: A survey of results and
  outlook.
\newblock {\em Proceedings of the IEEE}, 2021.

\bibitem{gale2019state}
Trevor Gale, Erich Elsen, and Sara Hooker.
\newblock The state of sparsity in deep neural networks.
\newblock {\em arXiv preprint arXiv:1902.09574}, 2019.

\bibitem{hafner2020mastering}
Danijar Hafner, Timothy Lillicrap, Mohammad Norouzi, and Jimmy Ba.
\newblock Mastering atari with discrete world models.
\newblock {\em arXiv preprint arXiv:2010.02193}, 2020.

\bibitem{han2019variational}
Kuan Han, Haiguang Wen, Junxing Shi, Kun-Han Lu, Yizhen Zhang, Di~Fu, and
  Zhongming Liu.
\newblock Variational autoencoder: An unsupervised model for encoding and
  decoding fmri activity in visual cortex.
\newblock {\em NeuroImage}, 198:125--136, 2019.

\bibitem{Herz2006}
Andreas~VM Herz, Tim Gollisch, Christian~K Machens, and Dieter Jaeger.
\newblock Modeling single-neuron dynamics and computations: a balance of detail
  and abstraction.
\newblock {\em science}, 314(5796):80--85, 2006.

\bibitem{hunsberger2016training}
Eric Hunsberger and Chris Eliasmith.
\newblock Training spiking deep networks for neuromorphic hardware.
\newblock {\em arXiv preprint arXiv:1611.05141}, 2016.

\bibitem{Illing2019}
Bernd Illing, Wulfram Gerstner, and Johanni Brea.
\newblock {Biologically plausible deep learning — But how far can we go with
  shallow networks?}
\newblock {\em Neural Networks}, 118:90--101, 2019.

\bibitem{kingma2014adam}
Diederik~P Kingma and Jimmy Ba.
\newblock Adam: A method for stochastic optimization.
\newblock {\em arXiv preprint arXiv:1412.6980}, 2014.

\bibitem{kingma2013auto}
Diederik~P Kingma and Max Welling.
\newblock Auto-encoding variational bayes.
\newblock {\em arXiv preprint arXiv:1312.6114}, 2013.

\bibitem{krizhevsky2012imagenet}
Alex Krizhevsky, Ilya Sutskever, and Geoffrey~E Hinton.
\newblock Imagenet classification with deep convolutional neural networks.
\newblock In {\em Advances in neural information processing systems}, pages
  1097--1105, 2012.

\bibitem{lecun1998gradient}
Yann LeCun, L{\'e}on Bottou, Yoshua Bengio, and Patrick Haffner.
\newblock Gradient-based learning applied to document recognition.
\newblock {\em Proceedings of the IEEE}, 86(11):2278--2324, 1998.

\bibitem{Lee2019}
Chankyu Lee, Syed~Shakib Sarwar, Priyadarshini Panda, Gopalakrishnan
  Srinivasan, and Kaushik Roy.
\newblock {Enabling Spike-based Backpropagation for Training Deep Neural
  Network Architectures}.
\newblock {\em Frontiers in Neuroscience}, 14, mar 2019.

\bibitem{Lee2016}
Jun~Haeng Lee, Tobi Delbruck, and Michael Pfeiffer.
\newblock {Training Deep Spiking Neural Networks Using Backpropagation}.
\newblock {\em Frontiers in Neuroscience}, 10(NOV):508, nov 2016.

\bibitem{Levy2004}
Ifat Levy, Uri Hasson, and Rafael Malach.
\newblock {One Picture Is Worth at Least a Million Neurons}.
\newblock {\em Current Biology}, 14(11):996--1001, jun 2004.

\bibitem{lu2019dying}
Lu~Lu, Yeonjong Shin, Yanhui Su, and George~Em Karniadakis.
\newblock Dying relu and initialization: Theory and numerical examples.
\newblock {\em arXiv preprint arXiv:1903.06733}, 2019.

\bibitem{mathieu2019}
Emile Mathieu, Tom Rainforth, N~Siddharth, and Yee~Whye Teh.
\newblock Disentangling disentanglement in variational autoencoders.
\newblock In Kamalika Chaudhuri and Ruslan Salakhutdinov, editors, {\em
  Proceedings of the 36th International Conference on Machine Learning},
  volume~97 of {\em Proceedings of Machine Learning Research}, pages
  4402--4412. PMLR, 09--15 Jun 2019.

\bibitem{Meo2021}
Cristian Meo and Pablo Lanillos.
\newblock Multimodal vae active inference controller.
\newblock {\em arXiv preprint arXiv:2103.04412}, 2021.

\bibitem{Neftci2019}
Emre~O. Neftci, Hesham Mostafa, and Friedemann Zenke.
\newblock {Surrogate Gradient Learning in Spiking Neural Networks: Bringing the
  Power of Gradient-based optimization to spiking neural networks}.
\newblock {\em IEEE Signal Processing Magazine}, 36(6):51--63, 2019.

\bibitem{OConnor2018}
Peter O'Connor, Efstratios Gavves, Matthias Reisser, and Max Welling.
\newblock {Temporally efficient deep learning with spikes}.
\newblock {\em 6th International Conference on Learning Representations, ICLR
  2018 - Conference Track Proceedings}, pages 1--19, 2018.

\bibitem{OConnor2016}
Peter O'Connor and Max Welling.
\newblock Deep spiking networks.
\newblock {\em arXiv preprint arXiv:1602.08323}, 2016.

\bibitem{orchard2015converting}
Garrick Orchard, Ajinkya Jayawant, Gregory~K Cohen, and Nitish Thakor.
\newblock Converting static image datasets to spiking neuromorphic datasets
  using saccades.
\newblock {\em Frontiers in neuroscience}, 9:437, 2015.

\bibitem{Panda2016}
Priyadarshini Panda and Kaushik Roy.
\newblock {Unsupervised regenerative learning of hierarchical features in
  Spiking Deep Networks for object recognition}.
\newblock In {\em Proceedings of the International Joint Conference on Neural
  Networks}, pages 299--306. Institute of Electrical and Electronics Engineers
  Inc., 2016.

\bibitem{Paulin2014}
Michael~G. Paulin and Andre van Schaik.
\newblock {Bayesian Inference with Spiking Neurons}.
\newblock {\em Encyclopedia of Computational Neuroscience}, pages 1--4, jun
  2014.

\bibitem{pfeiffer2018deep}
Michael Pfeiffer and Thomas Pfeil.
\newblock Deep learning with spiking neurons: opportunities and challenges.
\newblock {\em Frontiers in neuroscience}, 12:774, 2018.

\bibitem{qiao2019neural}
Siyuan Qiao, Zhe Lin, Jianming Zhang, and Alan~L Yuille.
\newblock Neural rejuvenation: Improving deep network training by enhancing
  computational resource utilization.
\newblock In {\em Proceedings of the IEEE/CVF Conference on Computer Vision and
  Pattern Recognition}, pages 61--71, 2019.

\bibitem{QuianQuiroga2010}
Rodrigo {Quian Quiroga} and Gabriel Kreiman.
\newblock {Measuring sparseness in the brain: Comment on Bowers (2009).}
\newblock {\em Psychological Review}, 117(1):291--297, 2010.

\bibitem{razavi2019generating}
Ali Razavi, Aaron van~den Oord, and Oriol Vinyals.
\newblock Generating diverse high-fidelity images with vq-vae-2.
\newblock {\em arXiv preprint arXiv:1906.00446}, 2019.

\bibitem{Roy2019}
Deboleena Roy, Priyadarshini Panda, and Kaushik Roy.
\newblock {Synthesizing images from spatio-temporal representations using
  spike-based backpropagation}.
\newblock {\em Frontiers in Neuroscience}, 13(JUN), 2019.

\bibitem{Schuman2017}
Catherine~D Schuman, Thomas~E Potok, Robert~M Patton, J~Douglas Birdwell,
  Mark~E Dean, Garrett~S Rose, and James~S Plank.
\newblock A survey of neuromorphic computing and neural networks in hardware.
\newblock {\em arXiv preprint arXiv:1705.06963}, 2017.

\bibitem{Tavanaei2016}
Amirhossein Tavanaei and Anthony~S Maida.
\newblock Bio-inspired spiking convolutional neural network using layer-wise
  sparse coding and stdp learning.
\newblock {\em arXiv preprint arXiv:1611.03000}, 2016.

\bibitem{Zambrano2019}
Davide Zambrano, Roeland Nusselder, H.~Steven Scholte, and Sander~M.
  Boht{\'{e}}.
\newblock {Sparse Computation in Adaptive Spiking Neural Networks}.
\newblock {\em Frontiers in Neuroscience}, 12(JAN):987, jan 2019.

\bibitem{Zenke2018}
Friedemann Zenke and Surya Ganguli.
\newblock {SuperSpike: Supervised Learning in Multilayer Spiking Neural
  Networks}.
\newblock {\em Neural Computation}, 30(6):1514--1541, jun 2018.

\bibitem{Zhang2019}
Malu Zhang, Hong Qu, Ammar Belatreche, Yi~Chen, and Zhang Yi.
\newblock {A Highly Effective and Robust Membrane Potential-Driven Supervised
  Learning Method for Spiking Neurons}.
\newblock {\em IEEE Transactions on Neural Networks and Learning Systems},
  30(1):123--137, 2019.

\end{thebibliography}

\vfill
\pagebreak

\appendix

\section{Appendix}
\label{sec:appendix}
\subsection{Training details}
\label{sec:training}

In this work, model weight parameters $\theta_l$ for all layers $l \in L$ are initialized randomly according to:
\begin{equation}
    \label{eq:weight_init}
    \theta_{l} \sim \mathcal{N}\left(0,\,\sqrt{\frac{2}{n_{l}}}\right)\,,
\end{equation}
where $n_l$ of is the input size in the fully connected case and kernel width $\times$ kernel height $\times$ number of input channels for convolutional and deconvolutional layers.

After initialization, model parameters are updated iteratively through backpropagation on batches of $N = 64$ training examples each for 10 epochs with a learning rate of $\alpha=0.0005$. The exact step size of each parameter value $\Delta\theta$ is determined by the ADAM optimization algorithm at standard settings \cite{kingma2014adam}.

\subsection{Networks \& Learning Parameters}
\label{sec:parameters}

%\todo{some explanation missing how the parameters were found.}

The key parameters describing the models and learning process are summarized in \autoref{tab:hyperparameters} and \autoref{tab:compared_models}. All parameter values were found in preliminary experiments through empirical testing. The reported default values are used throughout the reported experiments unless explicitly specified. 

The number of trainable parameters $\theta$ varies across models and scales linearly with layer 3 latent representation size $n_z$ according to \autoref{tab:parameters}.

\begin{table}[h!]
\centering
\caption[Trainable model parameters]{Trainable model parameters depend on the size of the latent representation. SAE models have fewer parameters compared to the baseline models due to the lack of bias parameters. VAE model parameters are increased because the number of neurons in layer three are effectively doubled by predicting both $\mu$ and $\log\sigma^2$, each with $n_z$ individual neurons.}
\begin{tabular}{rrrr}
\toprule
\textbf{latent size $z$} & \textbf{SAE}       & \textbf{AE}        & \textbf{VAE}       \\
\midrule
10              & 282,400   & 295,210   & 423,220   \\
20              & 538,400   & 551,220   & 807,240   \\
50              & 1,306,400 & 1,319,250 & 1,959,300 \\
\textbf{100}    & 2,586,400 & 2,599,300 & 3,879,400 \\
\bottomrule
\end{tabular}
\label{tab:parameters}
\end{table}

\begin{table}[h!]
    \centering
    \caption[Model hyper-parameters]{Overview of relevant hyper-parameters and their default values. The last three parameters are only relevant in the context of the SAE.}
    \begin{tabular}{l l l}
        \toprule
        & \textbf{Parameter} & \textbf{Default value}\\
        \midrule
        \textit{Architecture} & Kernel size & {5 x 5}\\
        & Convolution channels & {16, 32}\\
        & Latent size $n_z$ & 100\\
        
        \textit{Training} & Epochs & 10\\
        & Batch size $N$ & 64\\
        & Learning rate $\alpha$ & 0.0005\\
        
        \textit{Encoding} & Noise $\epsilon$ & 0.0\\
        & Scaling factor $s$ & 0.2\\
        
        \textit{Temporal dynamics} & Time steps $T$ & 100\\
        & Potential decay $\tau$ & 0.99\\
        \bottomrule
    \end{tabular}

    \label{tab:hyperparameters}
\end{table}

\paragraph{Autoencoder Parameters}

First, the baseline AE model was used to determine a suitable learning rate and L2 regularization parameter based on the lowest reconstruction error. The results of this search are summarized in \autoref{tab:ae_params}. The results show a large effect of the learning rate on the results with best and most reliable performance at $\alpha = 0.0005$. To enable fair model comparison, this learning rate was also chosen for all other models in all following experiments. Learning rates of $\alpha \ge 0.01$ resulted in the AE models to return all-zero image reconstructions, equivalent to a $\mathcal{L}_{rec}$ of approximately 89. Further, these results show that in all reported cases the AE models have at least 50\% dead neurons in layer 3, limiting each models information representation capacity in the latent space. The L2 regularization has a slight improvement on AE reconstruction error, with an optimum for $l_2=0.0001$. However, L2 regularization was not suitable for addressing the dying neuron problem.

\begin{table}[h!]
	\centering
	\caption[AE parameter search]{AE parameter search. First, a suitable learning rate was determined as $\alpha = 0.0005$ based on the lowest reconstruction loss without regularization. Later, L2 loss weight parameter was determined as $l_2=0.0001$. All runs were done over 10 epochs on the full dataset. Inactive neurons are reported for layer 3 only. Parameters chosen for further experiments are highlighted in bold. INP = inactive neuron proportion of layer 3.}
	\begin{tabular}{llrrrr}
	    \toprule
		\textbf{Parameter} & \textbf{Value}  & \multicolumn{2}{l}{\textbf{$\mathcal{L}_{rec}$}}  & \multicolumn{2}{l}{\textbf{INP}}         \\
		\midrule
		$\alpha$    & 0.000001  & 63.78 & $\pm$ 5.07 & 0.66 & $\pm$ 0.07 \\
		& 0.00001         & 30.14 & $\pm$ 12.89  & 0.67 & $\pm$ 0.22     \\
		& \textbf{0.0005} & \textbf{5.65} & \textbf{$\pm$ 0.94}  & \textbf{0.67} & $\pm$ \textbf{0.08} \\
		& 0.0001          & 15.68 & $\pm$ 9.86   & 0.72 & $\pm$ 0.15     \\
		& 0.001           & 45.86 & $\pm$ 43.61  & 0.65 & $\pm$ 0.30     \\
		& 0.01            & 89.46 & $\pm$ 0.00   & 0.91 & $\pm$ 0.04     \\
		& 0.1             & 89.46 & $\pm$ 0.00   & 0.90 & $\pm$ 0.04     \\
		\midrule
		$l_2$  & \textbf{0 } & \textbf{5.65} & \textbf{$\pm$ 0.94}  & \textbf{0.67} & $\pm$ \textbf{0.08} \\
		& 0.000001        & 5.41 & $\pm$ 1.41    & 0.66 & $\pm$ 0.10       \\
		& 0.00001         & 4.54 & $\pm$ 1.49    & 0.56 & $\pm$ 0.10       \\
		& \textbf{0.0001} & \textbf{4.50} & \textbf{$\pm$ 1.38}  & \textbf{0.62} & $\pm$ \textbf{0.11} \\
		& 0.001           & 5.57 & $\pm$ 1.89    & 0.72 & $\pm$ 0.10       \\
		& 0.01            & 6.54 & $\pm$ 1.18    & 0.78 & $\pm$ 0.04       \\
		& 0.1             & 7.98 & $\pm$ 1.25    & 0.81 & $\pm$ 0.05      \\
		\bottomrule
	\end{tabular}
	\label{tab:ae_params}
\end{table}

\paragraph{Variational Autoencoder Parameters}

As the next step, VAE regularization parameters were determined and the results of this search are summarized in \autoref{tab:vae_params}. In this case, first the weight of the Kullback–Leibler divergence term $\beta$ was fixed at its standard value of 1 to find a suitable L2 loss weight. In contrast to the AE model, for the VAE sufficient L2 regularization ($l_2\ge0.001$) was necessary to obtain any useful image reconstructions. The optimal parameter was found at $l_2=0.01$ and is therefore 100 times larger compared to the AE. 

In contrast, lower values for $\beta$ show a strong decrease in model reconstruction error. In fact, ignoring the Kullback–Leibler divergence term by setting $\beta = 0$ showed the lowest reconstruction error across \textit{all} experiments. However, in the case of low values for $\beta$, the $\sigma$ vector predicted by the encoder showed very small values (data not shown), rendering the "variation" quasi non-existent. This is not surprising, because the key function of this regularization term is to put a soft constraint on the values of $\sigma$ to be close to 1, so that $z$ is sampled with variance. The VAE with low values for $\beta$ can therefore be seen as a quasi-deterministic AE model. The difference in performance of these models (VAE $\beta=0$ $\mathcal{L}_{rec}$ = 1.646 $\pm$ 0.943) compared to the best AE configuration (AE $l_2=0.0001$ $\mathcal{L}_{rec}$ = 4.495 $\pm$ 1.382) can therefore be explained by the lack of a ReLU activation function on the predicted $\mu$ vector, and the resulting absence of dead neurons in layer 3. This is a strong indication that the dying neuron problem may have a large effect on model performance and should therefore be avoided as much as possible. Although the image reconstruction error is lowest in cases of low $\beta$, only the cases of $\beta = 1$ and $\beta= 0.1$ were considered for further analysis, as in these cases the variation is still functioning with values of $\sigma$ close to 1.

\begin{table}[h!]
	\centering
	\caption[VAE parameter search]{VAE parameter search. For comparability, the same learning rate as AE was used with $\alpha = 0.0005$. First, the L2 loss weight parameter was determined as $l_2=0.01$ with a fixed $\beta=1.0$. Later, the effect of different values for $\beta$ were tried with L2 regularization. All runs were done over 10 epochs on the full dataset. Parameters chosen for further experiments are highlighted in bold.}
	\begin{tabular}{llrr}
	    \toprule
		\textbf{Parameter} & \textbf{Value}  & \multicolumn{2}{l}{\textbf{$\mathcal{L}_{rec}$}}     \\
		\midrule
		$l_2$    & 0         & 89.46 & $\pm$ 0.00         \\
		& 0.000001        & 89.46 & $\pm$ 0.00         \\
		& 0.00001         & 89.46 & $\pm$ 0.00         \\
		& 0.0001          & 89.46 & $\pm$ 0.00         \\
		& 0.001           & 38.40 & $\pm$ 29.50        \\
		& \textbf{0.01}   & \textbf{19.14} & \textbf{$\pm$ 0.61}\\
		& 0.1             & 20.76 & $\pm$ 0.23         \\
		\midrule
		$\beta$  & 0      & 1.65 & $\pm$ 0.94          \\
		& 0.001           & 1.64 & $\pm$ 1.41          \\
		& 0.01            & 2.29 & $\pm$ 1.49          \\
		& \textbf{0.1}    & \textbf{5.94}    & \textbf{ $\pm$ 1.38} \\
		& \textbf{1}      & \textbf{19.14}    & \textbf{ $\pm$ 0.61}\\
		& 10              & 52.90 & $\pm$ 0.08         \\
		\bottomrule
	\end{tabular}
	\label{tab:vae_params}
\end{table}

\paragraph{Spiking Autoencoder Parameters}

For the SAE models, a scaling parameter $s$ is required for both information encoding and decoding. Specifically, this parameter directly influences the expected firing rate of neurons resulting from the Poisson encoding. A value of $s=1$ refers to a spiking probability of 100\% at each time step if the respective pixel value of an example image is 1 (assuming normalized pixel values). Similarly, the output layer would be expected to show the same constant bursting behavior for respective neurons to obtain a low reconstruction error. High firing rates may increase the precision with which the floating point pixel values are translated into sequences of discrete spike events. However, bursting at maximum firing rates is considered inefficient and aimed to be avoided with the regularization terms. It therefore seems not helpful to induce and expect bursting behavior, while at the same time punishing it. preliminary experiments showed that of all tested values for $0.1\le s \le 1$, results were most promising at $s = 0.2$ (data not shown). Note that this does not mean that the expected normalized average firing rate of neurons in any layer should be at 0.2. The MNIST dataset has a mean normalized pixel value of approximately 0.13 $\pm$ 0.31. Therefore, the expected normalized average firing rate of any Poisson encoded images (and ideally layer 6 output) is $0.2 * 0.13 = 0.026$. This matches well with the recorded average 0.025 (see green histogram in \autoref{fig:qualitative_results}D). For the purposes of this work, this is considered an overall sparse encoding.

Beyond the scaling parameter, we were looking for a set of regularization parameter values that that result low reconstruction error $\mathcal{L}_{rec}$ as well as a low percentage of inactive layer 3 neurons (INR), preferably 0. Preconceptions are different about the latent representation spike density, that is the average normalized firing rate of active neurons (AFR) only. Here, no exact optimum is desired. Instead spike density should neither be too high, indicating inefficient bursting behavior, nor should it be 0, indicating no activity whatsoever. Monitoring this variable in conjunction with reconstruction error is particularly interesting, as we find a trade-off between the total number of spike events and the precision with which information is encoded in the latent representation. In all cases, standard deviations for these outcome variables should be low, indicating robust effects of the regularization method.

Based on the desired effect of the different regularization types, introduced in \autoref{sec:regularization}, we expect potential regularization (P1 and P2) to prevent dead neurons (decrease percent inactive). Furthermore, activity regularization (A1) is expected to reduce the spike density, but may also increase the number of dead neurons when penalizing spiking behavior too strong. As this term penalizes every spike independent of the respective neurons' firing rate, we expect these rates to be generally lower compared to the unregularized case, but not necessarily evenly distributed across neurons. Finally, in order to both prevent dying neurons and bursting behavior, we expect that a combination of potential and activity regularization to be most suitable. 

SAE regularization methods were initially tested in preliminary experiments (1 epoch only) mostly independent of one another to find parameter values that can improve model performance. The results of this search are presented in \autoref{tab:sae_params_I}. The default network without any regularization performs poorly on the image reconstruction task with around 86\% of layer three neurons dead, while the remaining neurons engage in high firing rates. Weight regularization slightly improves the reconstruction loss and performance reliability, but instead increase the average spike density. The number of inactive neurons remains largely unchanged. Therefore, weight regularization seems unsuitable to address the problems of dead and bursting neurons in the SAE network. 

In contrast, potential regularization and activity regularization reduce reconstruction loss, spike density, and the number of inactive neurons for various parameter values. Interestingly, in all cases a parameter value of 0.01 for $p_1$, $p_2$, and $a_1$ shows the lowest reconstruction error. Stronger P1 and A1 regularization results in the network not learning the task, and weaker regularization results are comparable to the unregularized network. Although some parameter values are clearly preferable to others, in general potential regularization, specifically P2, seems suitable to avoid dead neurons as expected. %As expected, the combination of potential and activity regularization enables the combination of a reconstruction loss below any value activity regularization alone archived while maintaining a lower spike density compared to potential regularization by itself with the same parameter value.

\begin{table}[h!]
	\centering
	\caption[SAE Regularization parameter search I]{In this first phase of parameter search for the SAE, regularization methods were tested primarily independent of one another. Average firing rate of active neurons (AFR) and inactive neuron proportion (INP) are reported for layer 3 only. For these experiments, networks were trained for a single epoch to shorten overall runtime. Colors highlight high values in red and low values in green per column for easier visual comparison. }
		\begin{tabular}{llrrrrrr}
		\toprule
		\textbf{Parameter} & \textbf{Value} & \multicolumn{2}{l}{\textbf{$\mathcal{L}_{rec}$}}                & \multicolumn{2}{l}{\textbf{AFR}}                          & \multicolumn{2}{l}{\textbf{INP}}                          \\  \midrule
		None               &                & \cellcolor[HTML]{FFDF82}38.66 & \cellcolor[HTML]{F8696B}$\pm$ 10.19 & \cellcolor[HTML]{FECE7F}0.51 & \cellcolor[HTML]{FED781}$\pm$ 0.15 & \cellcolor[HTML]{FED780}0.86  & \cellcolor[HTML]{FDB87B}$\pm$ 0.18 \\  \midrule
		$l_2$              & 0.1            & \cellcolor[HTML]{FFE183}37.74 & \cellcolor[HTML]{FED380}$\pm$ 6.07  & \cellcolor[HTML]{F97A6F}0.64 & \cellcolor[HTML]{FFE383}$\pm$ 0.10       & \cellcolor[HTML]{FFE283}0.85  & \cellcolor[HTML]{FED781}$\pm$ 0.14 \\
		& 0.01           & \cellcolor[HTML]{FFE583}36.24 & \cellcolor[HTML]{FECB7E}$\pm$ 6.39  & \cellcolor[HTML]{FB9B75}0.59 & \cellcolor[HTML]{FFE683}$\pm$ 0.09       & \cellcolor[HTML]{FED380}0.87  & \cellcolor[HTML]{F0E683}$\pm$ 0.11 \\
		& 0.001          & \cellcolor[HTML]{FFE784}35.36 & \cellcolor[HTML]{FCA777}$\pm$ 7.80  & \cellcolor[HTML]{FCB179}0.56 & \cellcolor[HTML]{FFDF82}$\pm$ 0.12       & \cellcolor[HTML]{FFE583}0.85  & \cellcolor[HTML]{FFDC81}$\pm$ 0.14 \\
		& 0.0001         & \cellcolor[HTML]{FFEB84}33.57 & \cellcolor[HTML]{FFE082}$\pm$ 5.56  & \cellcolor[HTML]{FDBA7B}0.54 & \cellcolor[HTML]{FFE082}$\pm$ 0.11       & \cellcolor[HTML]{FFDD82}0.86  & \cellcolor[HTML]{FFEB84}$\pm$ 0.12 \\
		& 0.00001        & \cellcolor[HTML]{FFEB84}33.44 & \cellcolor[HTML]{F2E783}$\pm$ 4.72  & \cellcolor[HTML]{FDBE7C}0.53 & \cellcolor[HTML]{FFE784}$\pm$ 0.09       & \cellcolor[HTML]{FFE784}0.84  & \cellcolor[HTML]{FFE784}$\pm$ 0.13 \\  \midrule
		$p_1$              & 0.1            & \cellcolor[HTML]{F8696B}89.39 & \cellcolor[HTML]{63BE7B}$\pm$ 0.00  & \cellcolor[HTML]{63BE7B}0.00 & \cellcolor[HTML]{63BE7B}$\pm$ 0.00       & \cellcolor[HTML]{F8696B}1.00  & \cellcolor[HTML]{63BE7B}$\pm$ 0.00 \\
		& 0.01           & \cellcolor[HTML]{A9D27F}25.19 & \cellcolor[HTML]{88C87D}$\pm$ 1.23  & \cellcolor[HTML]{69BF7B}0.02 & \cellcolor[HTML]{6EC17B}$\pm$ 0.01       & \cellcolor[HTML]{6FC17B}0.07  & \cellcolor[HTML]{B1D47F}$\pm$ 0.06 \\
		& 0.001          & \cellcolor[HTML]{C6DA80}27.95 & \cellcolor[HTML]{FED680}$\pm$ 5.97  & \cellcolor[HTML]{8CCA7D}0.13 & \cellcolor[HTML]{AED37F}$\pm$ 0.03       & \cellcolor[HTML]{DDE182}0.66  & \cellcolor[HTML]{9CCE7E}$\pm$ 0.04 \\
		& 0.0001         & \cellcolor[HTML]{FFEA84}33.89 & \cellcolor[HTML]{FFDE82}$\pm$ 5.64  & \cellcolor[HTML]{FECC7E}0.51 & \cellcolor[HTML]{FB9E76}$\pm$ 0.38       & \cellcolor[HTML]{F8E983}0.80  & \cellcolor[HTML]{FFD981}$\pm$ 0.14 \\
		& 0.00001        & \cellcolor[HTML]{E1E282}30.51 & \cellcolor[HTML]{B8D67F}$\pm$ 2.81  & \cellcolor[HTML]{F8696B}0.67 & \cellcolor[HTML]{F8696B}$\pm$ 0.59       & \cellcolor[HTML]{F2E783}0.77  & \cellcolor[HTML]{FDC27D}$\pm$ 0.16 \\  \midrule
		$p_2$              & 0.1            & \cellcolor[HTML]{8ECA7D}22.71 & \cellcolor[HTML]{88C87D}$\pm$ 1.23  & \cellcolor[HTML]{71C27B}0.04 & \cellcolor[HTML]{68BF7B}$\pm$ 0.00       & \cellcolor[HTML]{63BE7B}0.00  & \cellcolor[HTML]{63BE7B}$\pm$ 0.00 \\
		& 0.01           & \cellcolor[HTML]{63BE7B}18.55 & \cellcolor[HTML]{C4DA80}$\pm$ 3.21  & \cellcolor[HTML]{7BC47C}0.07 & \cellcolor[HTML]{85C77C}$\pm$ 0.02       & \cellcolor[HTML]{63BE7B}0.00  & \cellcolor[HTML]{68BF7B}$\pm$ 0.00 \\
		& 0.005          & \cellcolor[HTML]{6EC17B}19.61 & \cellcolor[HTML]{C9DB80}$\pm$ 3.35  & \cellcolor[HTML]{86C87D}0.11 & \cellcolor[HTML]{9ACE7E}$\pm$ 0.02       & \cellcolor[HTML]{8ECA7D}0.23  & \cellcolor[HTML]{BCD780}$\pm$ 0.07 \\
		& 0.001          & \cellcolor[HTML]{9BCE7E}23.86 & \cellcolor[HTML]{FFEA84}$\pm$ 5.18  & \cellcolor[HTML]{7DC57C}0.08 & \cellcolor[HTML]{84C77C}$\pm$ 0.02       & \cellcolor[HTML]{8DCA7D}0.23  & \cellcolor[HTML]{C4DA80}$\pm$ 0.08 \\
		& 0.0001         & \cellcolor[HTML]{CFDD81}28.82 & \cellcolor[HTML]{F8E983}$\pm$ 4.91  & \cellcolor[HTML]{B3D57F}0.24 & \cellcolor[HTML]{FFE082}$\pm$ 0.11       & \cellcolor[HTML]{C7DA80}0.54  & \cellcolor[HTML]{FDC37D}$\pm$ 0.16 \\
		& 0.00001        & \cellcolor[HTML]{EDE582}31.65 & \cellcolor[HTML]{F96C6C}$\pm$ 10.07 & \cellcolor[HTML]{FAE983}0.45 & \cellcolor[HTML]{FFDE82}$\pm$ 0.12       & \cellcolor[HTML]{F7E883}0.80  & \cellcolor[HTML]{FFE784}$\pm$ 0.13 \\  \midrule
		$a_1$              & 0.1            & \cellcolor[HTML]{FDB67A}56.21 & \cellcolor[HTML]{EEE683}$\pm$ 4.57  & \cellcolor[HTML]{63BE7B}0.00 & \cellcolor[HTML]{63BE7B}$\pm$ 0.00       & \cellcolor[HTML]{63BE7B}0.00 & \cellcolor[HTML]{64BE7B}$\pm$ 0.00 \\
		& 0.01           & \cellcolor[HTML]{8AC97D}22.29 & \cellcolor[HTML]{BCD780}$\pm$ 2.95  & \cellcolor[HTML]{74C27B}0.05 & \cellcolor[HTML]{A1CF7E}$\pm$ 0.03       & \cellcolor[HTML]{81C67C}0.17  & \cellcolor[HTML]{FDC27D}$\pm$ 0.16 \\
		& 0.001          & \cellcolor[HTML]{DEE182}30.28 & \cellcolor[HTML]{FFDA81}$\pm$ 5.81  & \cellcolor[HTML]{F5E883}0.44 & \cellcolor[HTML]{FFE784}$\pm$ 0.09       & \cellcolor[HTML]{FDC07C}0.89  & \cellcolor[HTML]{97CD7E}$\pm$ 0.04 \\
		& 0.0001         & \cellcolor[HTML]{FFE784}35.13 & \cellcolor[HTML]{FCB079}$\pm$ 7.45  & \cellcolor[HTML]{FDBC7B}0.54 & \cellcolor[HTML]{FFEB84}$\pm$ 0.07       & \cellcolor[HTML]{FFE082}0.85  & \cellcolor[HTML]{FFE383}$\pm$ 0.13 \\
		& 0.00001        & \cellcolor[HTML]{FFEA84}33.77 & \cellcolor[HTML]{EEE683}$\pm$ 4.58  & \cellcolor[HTML]{FED480}0.50 & \cellcolor[HTML]{C9DB80}$\pm$ 0.05       & \cellcolor[HTML]{FFDD82}0.86  & \cellcolor[HTML]{FDEA83}$\pm$ 0.12 \\  \bottomrule
	\end{tabular}
\label{tab:sae_params_I}
\end{table}

After these promising initial results for potential and activity regularization methods in preliminary experiments, several combinations of loss terms were evaluated in longer experiments (10 epochs). The results are summarized in \autoref{tab:sae_params_II}. Interestingly, the unregularized model shows about the same reconstruction loss and number of dead neurons after 10 epochs of training as after a single epoch, but the average spike density of surviving neurons increased from 0.51 $\pm$ 0.15 to 0.75 $\pm$ 0.11. In contrast, all tested combinations of potential and activity regularization result in the SAE network learning the image reconstruction task successfully. Further, dead neurons can be reduced reliably and avoided all together in many cases.

As expected, a trade-off between spike density and reconstruction error is clearly visible in \autoref{tab:sae_params_II}. Generally speaking, models that show the lowest reconstruction errors (below 7) have a higher spike density (around 0.2) than models with slightly higher reconstruction error (between 10 and 12). This trade-off is found again in later experiments (see \autoref{fig:comparison}D \& \autoref{fig:comparison2}C). 

Interestingly, layer 3 spike density is reduced more effectively by assigning this cost to spikes emitted in all layer, while A1(l=3) paired with P2 results in the overall lowest reconstruction error for compared spiking models.

Based on these findings, three SAE models are selected for further analysis. The first model is the unregularized SAE as an additional baseline. Second choice is a model configuration for sparse latent representations based on potential and A1 regularization, hereafter referred to as \textit{sparse SAE}. Finally, a model from the other end of the error-sparsity trade-off is selected with potential and A1(l=3) regularization, hereafter referred to as \textit{dense SAE}. The parameter values for all three selected SAE models are summarized in \autoref{tab:compared_models}.

\begin{table}[h!]
	\centering
	\caption[SAE regularization parameter search II]{In this second phase of parameter search for the SAE, regularization methods were tested primarily combined of one another. Average firing rate of active neurons (AFR) and inactive neuron proportion (INP) are reported for layer 3 only. In these tests L1 regularization is excluded. All runs were done over 10 epochs on the full dataset. Parameters chosen for further experiments are highlighted in bold. All runs were done over 10 epochs on the full dataset. Parameters chosen for further experiments are highlighted in bold. Colors highlight high values in red and low values in green per column for easier visual comparison.}
	\begin{tabular}{lllllrrrrrr}
	    \toprule
		$l_2$     & $p_1$          & $p_2$          & $a_1$     & $a_{1,l=3}$        & \multicolumn{2}{l}{$\mathcal{L}_{rec}$}                                          & \multicolumn{2}{l}{AFR}                                               & \multicolumn{2}{l}{INP}                                                    \\ \midrule
		\textbf{} & \textbf{}      & \textbf{}     & \textbf{} & \textbf{} & \cellcolor[HTML]{F9706D}\textbf{38.09} & \cellcolor[HTML]{FA8E73}$\pm$ \textbf{8.96} & \cellcolor[HTML]{F96E6C}\textbf{0.75} & \cellcolor[HTML]{FB9F76}$\pm$ \textbf{0.11} & \cellcolor[HTML]{F96D6C}\textbf{0.83} & \cellcolor[HTML]{F8696B}$\pm$ \textbf{0.20} \\
		0.01      &                &               &           &           & \cellcolor[HTML]{F9756E}37.03          & \cellcolor[HTML]{FB9073}$\pm$ 8.85          & \cellcolor[HTML]{F97A6F}0.69          & \cellcolor[HTML]{FB9C75}$\pm$ 0.11          & \cellcolor[HTML]{F8696B}0.86          & \cellcolor[HTML]{F9786E}$\pm$ 0.18          \\
		0.001     &                &               &           &           & \cellcolor[HTML]{F8696B}39.63          & \cellcolor[HTML]{FA8170}$\pm$ 10.14         & \cellcolor[HTML]{F8696B}0.77          & \cellcolor[HTML]{F96E6C}$\pm$ 0.16          & \cellcolor[HTML]{F96A6C}0.85          & \cellcolor[HTML]{F96A6C}$\pm$ 0.20          \\
		0.0001    &                &               &           &           & \cellcolor[HTML]{F96D6C}38.89          & \cellcolor[HTML]{F97A6F}$\pm$ 10.71         & \cellcolor[HTML]{F96D6C}0.75          & \cellcolor[HTML]{F8696B}$\pm$ 0.17          & \cellcolor[HTML]{F8696B}0.86          & \cellcolor[HTML]{FA7E6F}$\pm$ 0.17          \\
		0.00001   &                &               &           &           & \cellcolor[HTML]{F9786E}36.29          & \cellcolor[HTML]{FCA777}$\pm$ 6.78          & \cellcolor[HTML]{F97A6F}0.69          & \cellcolor[HTML]{FA8471}$\pm$ 0.14          & \cellcolor[HTML]{F96C6C}0.84          & \cellcolor[HTML]{F9796E}$\pm$ 0.17          \\ \midrule
		& 0.01           &         &           &           & \cellcolor[HTML]{FFDB81}12.92          & \cellcolor[HTML]{FDBE7C}$\pm$ 4.76          & \cellcolor[HTML]{C1D980}0.09          & \cellcolor[HTML]{FDBC7B}$\pm$ 0.07          & \cellcolor[HTML]{FFE383}0.06          & \cellcolor[HTML]{FEC97E}$\pm$ 0.06          \\
		& 0.005          & 0.005               &           &           & \cellcolor[HTML]{CEDD81}8.27           & \cellcolor[HTML]{FFE583}$\pm$ 1.33          & \cellcolor[HTML]{FFE884}0.15          & \cellcolor[HTML]{FED280}$\pm$ 0.04          & \cellcolor[HTML]{63BE7B}0.00          & \cellcolor[HTML]{63BE7B}$\pm$ 0.00          \\
		\textbf{} & \textbf{0.005} & \textbf{0.005} & \textbf{0.01} & \textbf{} & \cellcolor[HTML]{FFE082}\textbf{11.58} & \cellcolor[HTML]{FFEA84}$\pm$ \textbf{0.89} & \cellcolor[HTML]{93CB7D}\textbf{0.06} & \cellcolor[HTML]{FFEA84}$\pm$ \textbf{0.02} & \cellcolor[HTML]{FFE182}\textbf{0.07} & \cellcolor[HTML]{FED17F}$\pm$ \textbf{0.04} \\
		& 0.005          & 0.005          &          & 0.01     & \cellcolor[HTML]{B3D57F}7.88           & \cellcolor[HTML]{FFE984}$\pm$ 1.05          & \cellcolor[HTML]{FFEA84}0.15          & \cellcolor[HTML]{FED07F}$\pm$ 0.05          & \cellcolor[HTML]{63BE7B}0.00          & \cellcolor[HTML]{63BE7B}$\pm$ 0.00          \\ \midrule
		&                & 0.01           &                   &           & \cellcolor[HTML]{B1D47F}7.85           & \cellcolor[HTML]{FFDF82}$\pm$ 1.92          & \cellcolor[HTML]{FFDA81}0.22          & \cellcolor[HTML]{C5DA80}$\pm$ 0.01          & \cellcolor[HTML]{63BE7B}0.00          & \cellcolor[HTML]{63BE7B}$\pm$ 0.00          \\
		&                & 0.01           & 0.01              &           & \cellcolor[HTML]{FFEB84}9.08           & \cellcolor[HTML]{A6D17E}$\pm$ 0.44          & \cellcolor[HTML]{D5DF81}0.11          & \cellcolor[HTML]{DEE182}$\pm$ 0.01          & \cellcolor[HTML]{FFE984}0.02          & \cellcolor[HTML]{FFE784}$\pm$ 0.01          \\
		&                & 0.01           &          & 10               & \cellcolor[HTML]{FED280}15.07          & \cellcolor[HTML]{FFE583}$\pm$ 1.39          & \cellcolor[HTML]{74C27B}0.04          & \cellcolor[HTML]{67BF7B}$\pm$ 0.00          & \cellcolor[HTML]{FED680}0.15         & \cellcolor[HTML]{FED480}$\pm$ 0.04          \\
		&                & 0.01           &          & 1                & \cellcolor[HTML]{DAE081}8.43           & \cellcolor[HTML]{B4D57F}$\pm$ 0.49          & \cellcolor[HTML]{D5DE81}0.11          & \cellcolor[HTML]{FFEB84}$\pm$ 0.01          & \cellcolor[HTML]{99CD7E}0.00          & \cellcolor[HTML]{ABD27F}$\pm$ 0.00          \\
		&                & 0.01           &          & 0.1                & \cellcolor[HTML]{71C27B}6.94           & \cellcolor[HTML]{65BE7B}$\pm$ 0.18          & \cellcolor[HTML]{FFDF82}0.20          & \cellcolor[HTML]{F2E783}$\pm$ 0.01          & \cellcolor[HTML]{63BE7B}0.00          & \cellcolor[HTML]{63BE7B}$\pm$ 0.00          \\
		\textbf{} & \textbf{}      & \textbf{0.01}  & \textbf{}   & \textbf{0.01}  & \cellcolor[HTML]{64BE7B}\textbf{6.75}  & \cellcolor[HTML]{79C47C}$\pm$ \textbf{0.26} & \cellcolor[HTML]{FFDC81}\textbf{0.21} & \cellcolor[HTML]{FFE884}$\pm$ \textbf{0.02} & \cellcolor[HTML]{63BE7B}\textbf{0.00} & \cellcolor[HTML]{63BE7B}$\pm$ \textbf{0.00} \\
		\bottomrule
	\end{tabular}
\label{tab:sae_params_II}
\end{table}

\subsection{Software \& Hardware Implementation}

% software
All code used for this project was written in Python 3.7 and is available on github\footnote{\hyperlink{https://github.com/jhuebotter/SpikingVAE}{https://github.com/jhuebotter/SpikingVAE}}.%\footnote{\hyperlink{https://github.com/jhuebotter/SpikingVAE}{https://github.com/jhuebotter/SpikingVAE}}.
For deep learning the pytorch framework (version 1.5) was extended with several custom classes to accommodate the LIF functionality. Result images are generated with matplotlib (version 3.1.3) and seaborn (version 0.11) and statistical tests were conducted using scipy (version 1.5.2). Weights \& Biases were used for experiment tracking \cite{wandb}.

% hardware
The training, validation, and testing of all models was performed on RTX 2080Ti GPUs (Nvidia, USA) as part of the Distributed ASCI Supercomputer 5 (DAS-5) server \cite{bal2016medium}.

\subsection{MNIST Dataset}

The dataset chosen is the Modified National Institute of Standards and Technology (MNIST) database \cite{lecun1998gradient}. The dataset contains 28 x 28 gray-scale images and labels of handwritten digits from 0 to 9 with 60.000 examples for training and 10.000 for testing. All images are scaled in the range [0, 1] by default and consist only of a single channel. MNIST has been utilized as a well studied benchmark in computer vision tasks for the past two decades, and therefore is an easy first choice for investigating performance of novel neural networks against established methods. 

Typically, SNNs receive continuous streams of time-varying stimuli, which MNIST images in their default representation are not. Although the images can be and are encoded in a way that includes the dimension of time as described in \autoref{eq:poisson_encoding}, the underlying source of the stimuli is still static. Arguably, SNNs may not be capable of utilizing their full potential in such use cases when compared to their non-spiking counterparts. Alternative approaches do exist, such as Neuromorphic-MNIST by \cite{orchard2015converting}, who converted the original dataset examples into a temporal series of spikes using event-based cameras and saccade-like movements across the input space. This dataset was not used in this work, however, to maintain comparability between the novel SAE and baseline models in the input reconstruction task and remains a possible extension for further research.

\subsubsection{Information Bottlenecks}
\label{sec:information_bottlenecks}

\begin{figure}[h!]
    \centering
    \includegraphics[width=0.85\textwidth]{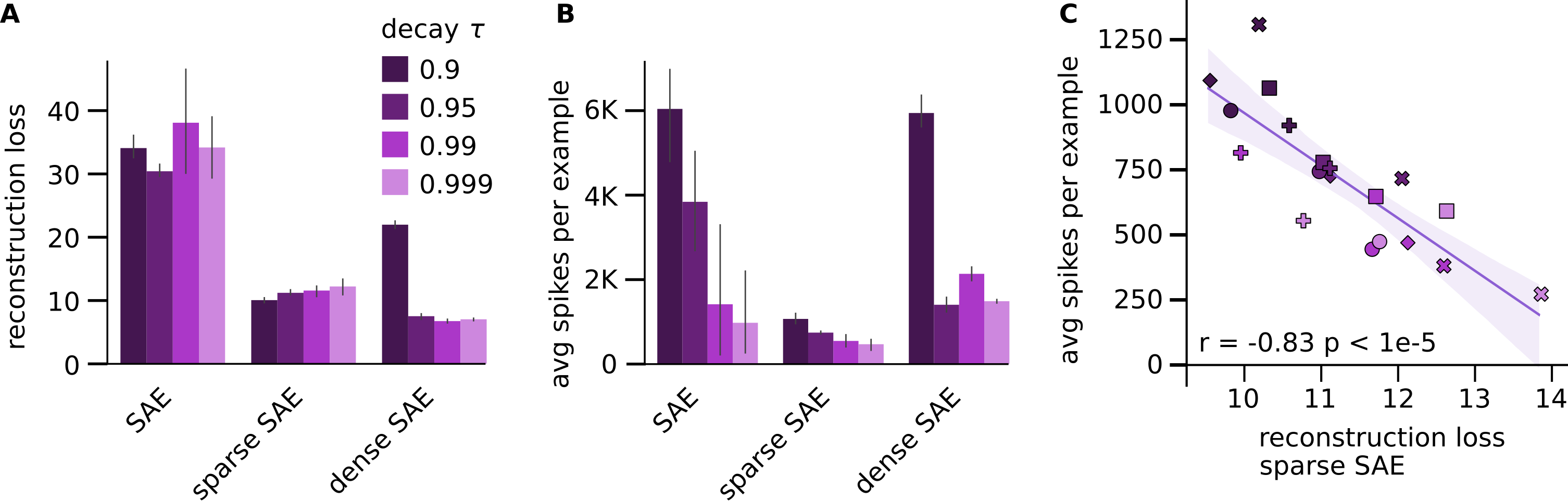}
    \caption[Quantitative model comparison]{\textbf{A} Changes in the membrane potential decay $\tau$ effects SAE models' reconstruction performance differently. \textbf{B} Increasing the decay variable generally leads to a reduction in avg. spikes per example across spiking models. \textbf{C} Sparse SAE model error shows significant negative correlation with reconstruction error with visible differences across randomly initialized repetitions.}
    \label{fig:comparison2}
\end{figure}

For the following experiments the models are trained and tested under different  information bottleneck conditions. Each time the value of a single variable is changed from its default value (see \autoref{tab:hyperparameters}). A visual summary of key findings is described in \autoref{fig:comparison} and \autoref{fig:comparison2}.

Reducing the size of the latent representation $n_z$ generally increases model error, except for VAE (\autoref{fig:comparison}A). Both AE models fail to reliably solve the reconstruction task with $n_z \leq 50$. The degree by which the SAE model performance seems affected by the reduced latent size seems to lie between the AE model (drastic change in model performance) and VAE (small change in model performance).

As expected, increasing the noise $\epsilon$ in the input signal generally increases reconstruction error across models (\autoref{fig:comparison}C). No tested model was capable of solving the reconstruction task with $\epsilon=0.9$. In contrast to the spiking models, AE$_{l2}$ and both VAE models show no decline in performance for low levels of noise with $\epsilon=0.3$. However, compared to the AE models, SAE models show lower variance in reconstruction error with $\epsilon=0.6$.

As expected, decreasing the number of simulated time steps $T$ in SAE models generally increases reconstruction error (\autoref{fig:comparison}B). In contrast to the dense SAE, the sparse model cannot solve the task with $T=10$. Dense SAE models reliably show a trade-off between the number of avg. spikes per example and the reconstruction error when simulated step number is varied (\autoref{fig:comparison}D). While a linear regression shows a significant negative correlation between both variables ($r=-0.83$, $p<1E-5$), visual inspection shows that this relationship is likely not linear, but rather exponential.

Changes in the membrane potential decay $\tau$ effects SAE models' reconstruction performance differently (\autoref{fig:comparison2}A). While the sparse SAE shows optimal performance for $\tau=0.9$, the dense model performs worst in this case and best with $\tau=0.99$. A possible explanation for the latter could be that the optimization of other model parameters was conducted with the default decay of $\tau=0.99$. This, however, does not explain why the trend of model performance is reversed from a sparse to a dense model. As expected, increasing the decay variable generally leads to a reduction in avg. spikes per example across spiking models (\autoref{fig:comparison2}B), with some exceptions for the dense model. Interestingly, sparse SAE model error shows significant negative correlation with reconstruction error ($r=-0.83$, $p<1E-5$) with visible influence of random initialization across repetitions (\autoref{fig:comparison2}C).

In summary, model behavior is affected by information bottlenecks mostly as expected. The regularized SAE models are somewhat robust to added noise in the input, which was as also reported for SNNs applied to MNIST classification \cite{Zhang2019, Tavanaei2016}. However, they show no clear advantage in this domain compared to non-spiking baseline models. In general, the spiking models' robustness towards changes in noise and latent size seems to lie above the AE models, but below VAE models.

\begin{comment}
\begin{itemize}
    \item Model reconstruction with noise
    \item Model reconstruction with smaller latent space
    \item Model reconstruction with less time (SAE only)
    \item Model reconstruction with different decay (SAE only)
    \item Model reconstruction with latent perturbation (dropout and noise / jitter)?
\end{itemize}
\end{comment}

\subsection{Latent Representation Analysis}
\label{sec:latent_analysis}

In this section, the models' latent representations of MINST examples are inspected further. For this purpose, of all fully trained models with default parameters, the instances lowest image reconstruction error are chosen for processing a single batch of 100 examples from the validation set with 10 examples per class. 

All three compared models encode the observed data in a compressed latent representation $z$, but the nature of this representation is model-specific. For the AE model, the latent representation $z$ of $x$ is a vector of $n_z$ positive rational elements $z_i \in \{\mathbb{Q} \ge 0\}$. In the VAE, $z$ is sampled from the multivariate distribution parameterized by the encoder output $\mu$ and $log(\sigma^2)$ as $z \sim \mathcal{N}(\mu(x'),\,\sigma(x'))$. Therefore, for the VAE model, the latent representation $z$ of $x$ is a vector of $n_z$ rational elements $z_i \in \mathbb{Q}$. Finally, the SAE has a temporal dimension in its representation of $z$, so that it has the shape $n_z$ x $T$, but every element is either 0 or 1. As the SAE encoding and decoding methods are based on rate code, the latent representation of this model is summarized as the average firing rates per neuron in the following analysis to enable comparison with the baseline models. Latent representations are scaled in the range [0, 1] for AE and SAE models and in the range [-1, 1] for VAE models. %The analysis of latent representation in the following subsections is based on a visual inspection on this activity matrix representation.

\begin{comment}
\begin{itemize}
    \item Hierarchical clustering as unsupervised learning
    \item Encoding correlation
    \item Sparsity in time and space
\end{itemize}
\end{comment}

\subsubsection{Latent Representation Clustering}

\begin{figure}[h!]
    \centering
    \includegraphics[width=\textwidth]{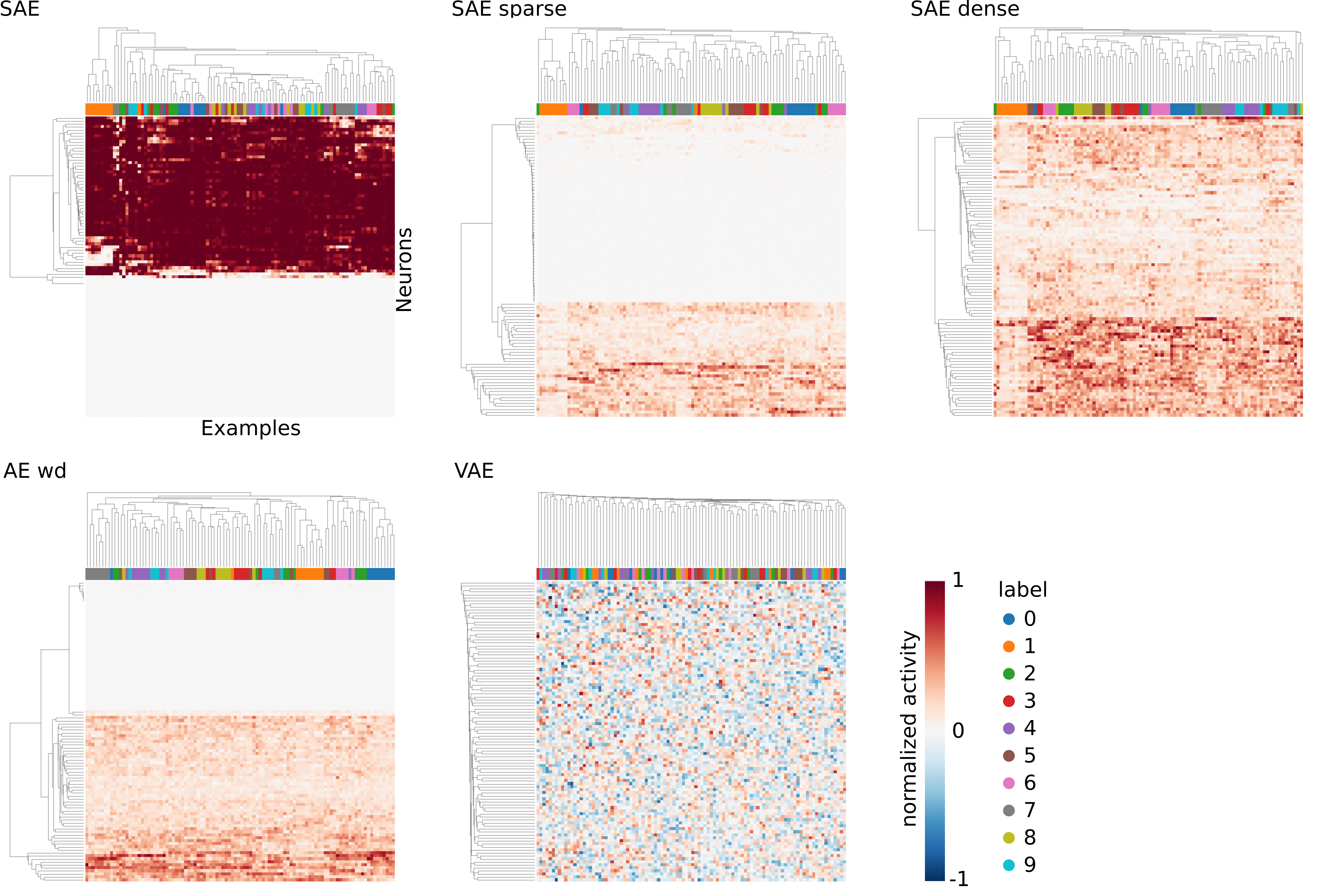}
    \caption[Latent representation clustering]{Comparison of the models' latent representation based on normalized activity. Regularization of SAE models prevents dead neurons as well as bursting behavior and activity patters can successfully be grouped by example labels using hierarchical clustering.}
    \label{fig:clustering}
\end{figure}

For the first comparison, the normalized layer 3 activity is represented in a matrix of 100 examples x 100 neurons for each model, a selection of which is shown in \autoref{fig:clustering}. Each column representing a single MNIST example is marked with a color representing that example's class label. Finally, an out-of-the-box hierarchical clustering algorithm is used to calculate linkage between both individual rows and columns based on their Euclidean distance and sort the matrix according to the resulting dendogram. Alternative distance metrics are evaluated in \autoref{tab:distances}. The resulting color band on top of each matrix can thereby give an indication, if examples of the same class are grouped together based on similar activity pattern in the latent representation of each model, without any model ever having seen the label data. 

The activity matrix of the unregularized SAE model (\autoref{fig:clustering} top left) clearly shows that approximately half of its layer 3 neurons are inactive across all examples, while the other half is engaged in bursting behavior most of the time. The sparse SAE model's activity matrix (\autoref{fig:clustering} top center) as expected, shows a drastically reduced average firing rate across neurons. It also shows that while technically there are no dead neurons, the majority of neurons actually exhibit very low firing rates consistently across examples. This clearly demonstrates that the regularization method prevents dead neurons and bursting behavior, but it does not at all guarantee that all neurons contribute to the representation of information equally with comparable firing rates. In contrast, the dense SAE model's activity matrix (\autoref{fig:clustering} top right) shows that bursting and dead neurons are prevented successfully. In this case, regularization enables a more balanced distribution of activity across neurons compared to both other spiking models with an average firing rate comparable to that of the more active group of neurons of the sparse SAE. The AE$_{l2}$ model's matrix (\autoref{fig:clustering} bottom left) shows a substantial amount of dead neurons. This suggests that L2 regularization is not a suitable method to prevent the dying ReLU problem, which makes sense . In contrast, the VAE (\autoref{fig:clustering} bottom center) exhibits no dead neurons and very balanced yet independent (non-similar) patterns of neuronal activity across examples. This behavior is expected because this model's layer 3 has no ReLU activation function and the KL-divergence regularization term is an attempt to have neuron activity distributed $\sim \mathcal{N}(0, 1)$. 

The clustering of examples shows, for the unregularized SAE, most examples of digit 1 are next to each other and some smaller clusters of 7s and 0s exist, but generally the ability of the clustering algorithm to find semantically meaningful structure in the activity matrix is limited. Interestingly, the color bar of both regularized SAE models show more structure in the latent representation, with large groups of the same color clustered together. A similar structure can be found in the AE$_{l2}$ model. For the VAE, however, hierarchical clustering is not able to identify any structure in the latent representation that suggests the class of each can be identified based on similar neuron activity patterns. 

\begin{comment}
\begin{itemize}
    \item Hierarchical clustering of examples based on avg. neuron activity
    \item Correlation of avg. neuron activity across examples?
    \item Correlation of example representation across neurons avg. activity?
\end{itemize}
\end{comment}

\begin{table}[h!]
\caption{The average distances between latent representations of MNIST examples within the same class $\overline{s}_{intra}$ are expected to be smaller than to examples of a different stimulus category $\overline{s}_{inter}$, if structural similarity is maintained in the learned representation. The ratios of $\overline{s}_{intra} / \overline{s}_{inter}$ suggest that this is the case for SAE and AE models, but not for VAE across several distance metrics.}
\begin{tabular}{llllll}
\toprule
           & euclidean       & stand. euclidean & squared euclidean & manhattan       & correlation     \\
\midrule
SAE        & 0.65 $\pm$ 0.07 & 0.69 $\pm$ 0.10  & 0.51 $\pm$ 0.06   & 0.62 $\pm$ 0.06 & 0.52 $\pm$ 0.05 \\
SAE sparse & 0.74 $\pm$ 0.01 & 0.86 $\pm$ 0.02  & 0.59 $\pm$ 0.01   & 0.75 $\pm$ 0.01 & 0.58 $\pm$ 0.01 \\
SAE dense  & 0.78 $\pm$ 0.00 & 0.80 $\pm$ 0.00  & 0.64 $\pm$ 0.00   & 0.77 $\pm$ 0.00 & 0.66 $\pm$0.01  \\
AE         & 0.65 $\pm$ 0.08 & 0.70 $\pm$ 0.07  & 0.46 $\pm$ 0.10   & 0.66 $\pm$ 0.08 & 0.46 $\pm$ 0.11 \\
AE$_{l2}$  & 0.69 $\pm$ 0.04 & 0.80 $\pm$ 0.03  & 0.52 $\pm$ 0.05   & 0.73 $\pm$ 0.03  & 0.53 $\pm$ 0.05 \\
VAE        & 0.97 $\pm$ 0.00 & 0.98 $\pm$ 0.00  & 0.95 $\pm$ 0.00   & 0.97 $\pm$ 0.00 & 0.95 $\pm$ 0.00 \\
$\beta$VAE & 0.94 $\pm$ 0.00 & 0.95 $\pm$ 0.00  & 0.88 $\pm$ 0.00   & 0.93 $\pm$ 0.00 & 0.88 $\pm$ 0.00 \\
\bottomrule
\end{tabular}
\label{tab:distances}
\end{table}

\subsubsection{Example Activity Correlation}

In this section, the column values (examples) of the normalized activity matrices from \autoref{fig:clustering} are correlated with one another for each model. This allows identification of clusters of examples with similar neuronal representation. 

\begin{figure}[h!]
    \centering
    \includegraphics[width=\textwidth]{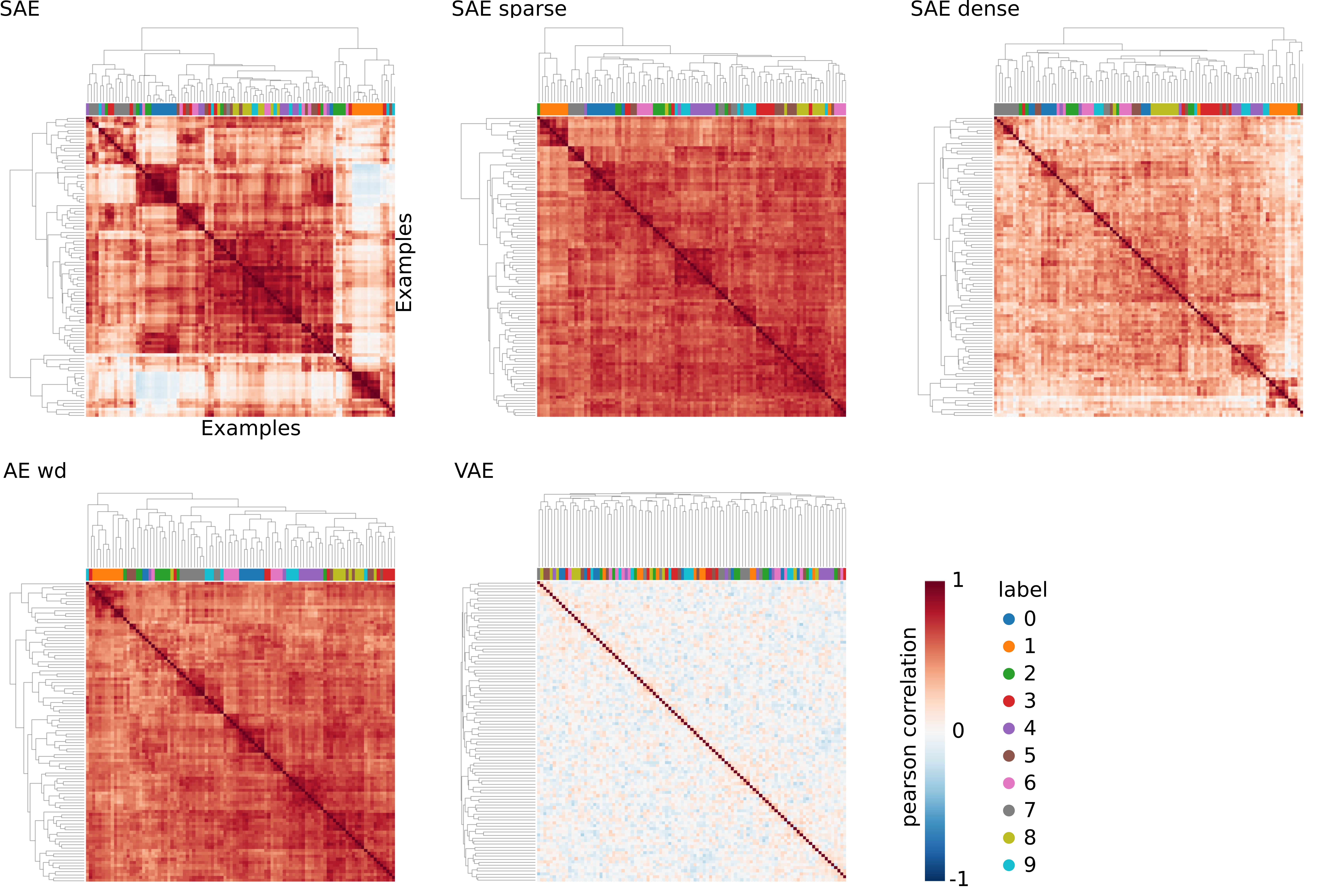}
    \caption[Example activity correlation]{Comparison of the example activity correlation of each models' latent representation with hierarchical clustering.}
    \label{fig:example_clustering}
\end{figure}

The SAE example cluster map (\autoref{fig:example_clustering} top left) shows that many example representation pairs have a strong positive correlation and a moderate negative correlation in between images of the classes 0 and 1. Regularization of the spiking model results in a more uniform positive correlation between examples with no negative correlations (\autoref{fig:example_clustering} top center and top right). The dense SAE, however, has a clearly reduced average correlation compared to both other spiking models. This is another indication that indeed activity is more balanced across neurons. The correlation matrix of the AE$_{l2}$ models (\autoref{fig:example_clustering} bottom left) strongly resembles that of the sparse SAE. This similarity in high overall example correlations is probably due to the clear trend for some neurons to be consistently more or less active across examples in both models. As expected from the results of the previous section, there is very little correlation between latent representation of examples encoded by the VAE model (\autoref{fig:example_clustering} bottom center). Further, the clustering of the example correlations show similar patterns across all models to those described by the preceding analysis. 

\subsubsection{Neuron Activity Correlation}

In this final part of the latent representation analysis each row's values (neurons) of the normalized activity matrices from \autoref{fig:clustering} are correlated with one another for each model. All neurons with no spikes in any example are excluded from this analysis. This allows identification of clusters of active neurons with similar activity patterns across examples. 

\begin{figure}[h!]
    \centering
    \includegraphics[width=\textwidth]{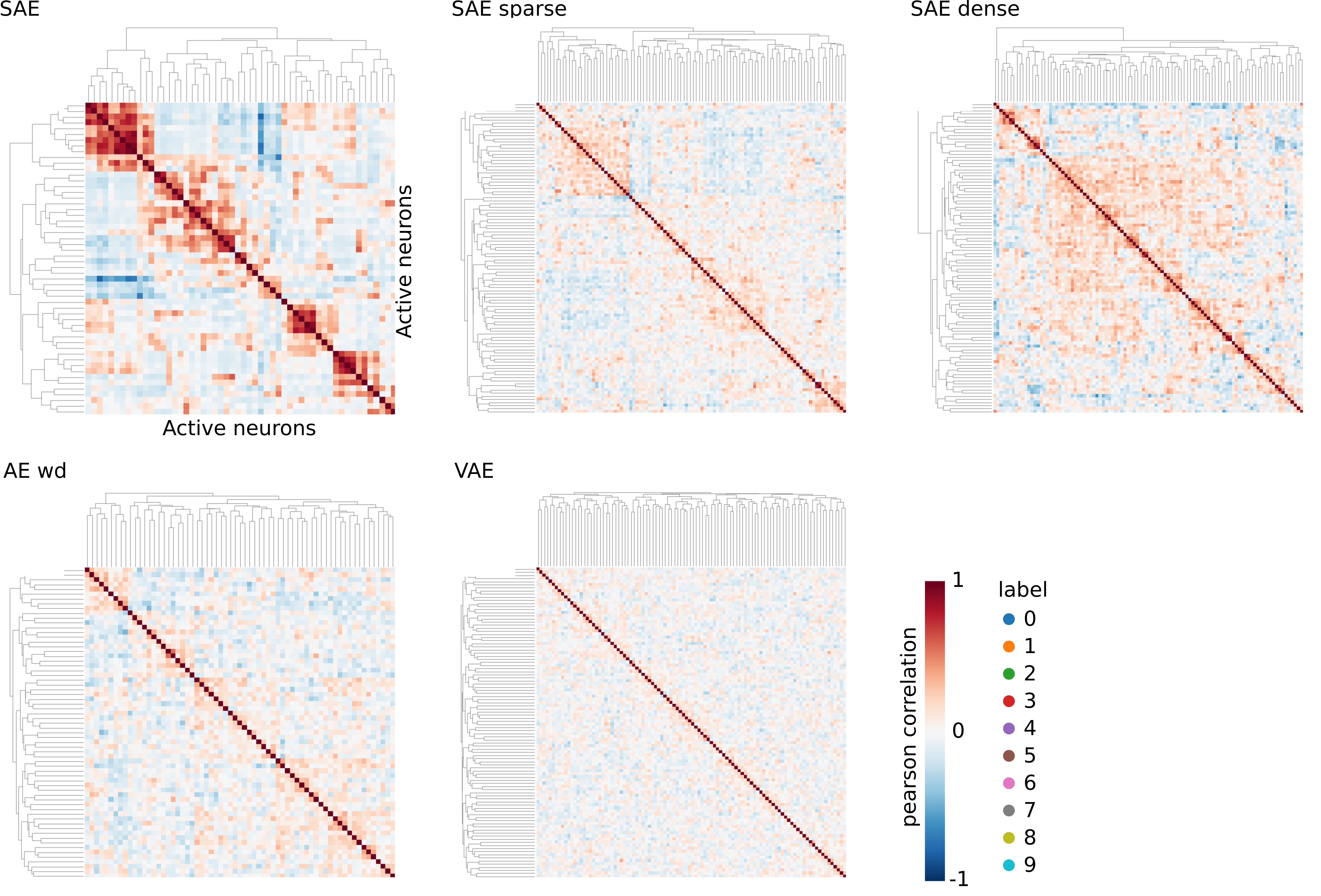}
    \caption[Neuron activity correlation]{Comparison of the neuronal activity correlation of each models' latent representation with hierarchical clustering.}
    \label{fig:neuron_clustering}
\end{figure}

The neuron activity cluster map of the unregularized SAE model (\autoref{fig:neuron_clustering} top left) shows the most prominent clusters of neurons with the strongest correlation coefficients (both positive and negative). The average absolute correlation and clustering is decreasing in the dense SAE (\autoref{fig:neuron_clustering} top lright), followed by the comparable sparse SAE (\autoref{fig:neuron_clustering} top center), AE$_{l2}$ (\autoref{fig:neuron_clustering} bottom left), and finally VAE (\autoref{fig:neuron_clustering} bottom center). The strong correlation in some models suggests that there is at least some redundant information encoded by different sub-populations of neurons in these models. This redundancy, however, does not seem to improve robustness towards noise in the input images. Finally, the described differences in latent representation qualities need to be considered when using a probabilistic VAE instead of a SAE as a model for unsupervised learning in the brain. \cite{Burbank2015} states that "a model which does not require neurons to be uncorrelated is desirable because neurons in real cortical networks respond to stimuli in highly correlated ways."

\subsection{Comparison to Earlier Approaches}
\label{sec:approach_comparison}

In this section I aim to relate this work to the two most relevant previous approaches to unsupervised learning in SAEs by \cite{Burbank2015} and \cite{Panda2016}. 

% network architecture
\paragraph{Network Architecture}
In her implementation, \cite{Burbank2015} uses two fully connected layers including feedback connections with LIF neurons with synaptic delay but otherwise unknown temporal dynamics. In contrast to this work, her input layer is also functioning as the output layer of the model. The hidden representation, however, is not smaller but larger than the input representation as her network has 392 input neurons and 5000 hidden neurons. In contrast to this work and \cite{Panda2016}, \cite{Burbank2015} follows Dale's principle and separates excitatory and inhibitory neurons into distinct populations.

\cite{Panda2016} use only convolutional and pooling layers in their SAE implementation. Slightly different networks are used for unsupervised feature learning on MNIST (1 hidden layer) and CIFAR-10 (2 hidden layers). Their kernel size is comparable to the 5x5 filters used in this project, although it is unclear if 5x5 (reported in text) or 7x7 (shown in images) was used. As their MNIST network has 16 and 64 learned filters in their first and second layer respectively, and no fully connected layer in the AE network, the imposed information bottleneck is much weaker compared to this work (16 and 32 filters + fully connected layer with up to 100 neurons in the encoder). %The LIF neurons used in this project have a refractory period with no-distinction between excitatory and inhibitory neurons.

% image reconstruction process
\paragraph{Image Processing}
\cite{Burbank2015} used a specific pre-processing of image data consisting of downscaling and subtraction of the dataset mean of each pixel before presenting images to the network as two separate non-negative "ON" and "OFF" maps. In agreement with this work, both \cite{Panda2016} and \cite{Burbank2015} used a Poisson encoding process to obtain spike sequences from images for network input.

\cite{Burbank2015} presented the input to her network for a short duration of ca. 10\,ms and in a second phase of similar duration without stimulus presentation input layer activity arising from feedback is recorded as network output. This short duration of stimulus presentation is likely a strong limiting factor of model performance. The output activity is substantially lower in spike counts than the input and requires an additional scaling parameter to be fitted before comparing network input to its output.\cite{Panda2016} showed each example to their network for 250\,ms. Unfortunately, neither \cite{Panda2016} nor \cite{Burbank2015} report how many discrete time steps were simulated in their experiments.

% learning rule
\paragraph{Learning Rule}
The learning rule developed by \cite{Burbank2015} is called mirrored STDP, because the temporal learning dynamics are reversed for excitatory feedforward and feedback connections. How this approach is extended to the inhibitory neurons in her network is not clear. This approach is deemed more biologically plausible than backpropagation, because it requires only local information for weight updates at individual synapses. After learning, network weights are symmetric, which is not necessary in the approach shown in this work.

\cite{Panda2016} used a regenerative learning approach based on error backpropagation on the membrane potential similar to this work. In contrast to my approach, however, \cite{Panda2016} trained each layer individually and in sequence. For this purpose, an additional pseudo-visible output layer was necessary. The error of this network was calculated based on the precise spike timing, requiring a backpropagation operation after every simulated time step. Why this is relevant for the reconstruction of static images with randomized spike generation process is not clear. 

In contrast to the presented approach, both \cite{Panda2016} and \cite{Burbank2015} used no mini-batches for training.

% regularization methods
\paragraph{Regularization Methods}
\cite{Burbank2015} used an additional synaptic scaling parameter for lifetime sparsity regularization of hidden neurons requiring a target activity hyper-parameter to be defined. The scaling parameter, although functionally identical with a typical bias, is not learned, but adjusted by a predefined rule in dependence on each neuron's recent average activity. This is a different attempt to a similar goal as the inclusion of the loss terms added in this work, although here target activity is not defined explicitly but implicitly. \cite{Panda2016} used no regularization method and no bias term.

\paragraph{Results}
Both \cite{Panda2016} and \cite{Burbank2015} evaluated their SAEs on MNIST and additionally CIFAR-10. \cite{Burbank2015} trained the fully connected network for two epochs, but the number of dataset sweeps (per layer) for \cite{Panda2016} is unknown. 

The lack of convolutional layers in Burbank's network requires each hidden neuron to learn a receptive field across the entire input space without position or size invariance. While this lead to fixed position Gabor-like filters during CIFAR-10 training, this was not the case for MNIST. The resulting image reconstructions are recognizable but of poor quality for both datasets. \cite{Burbank2015} evaluated image reconstruction by correlating the spike counts of the input/output layer from the stimulus presentation phase with the feedback phase. \cite{Panda2016} calculated their normalized MSE different to the standard deep learning approach used in this work. Unfortunately, numerical comparison between these different approaches is therefore not possible.

While image reconstruction on CIFAR-10 was blurry but promising for \cite{Panda2016}, \cite{Burbank2015} reports runaway network excitation which lead to non-interpretable results. In contrast to the spike regularization applied here, synaptic scaling failed to reliably avoid bursting behavior, although the loss term proposed here has yet to be evaluated on a network with feedback connections. 

Analysis of the learned hidden representation by \cite{Burbank2015} showed a distributed population code that neither requires nor enforces hidden unit decorrelation. This feature has been reproduced in this work, but with a different learning algorithm.

After the layer-wise unsupervised hierarchical feature learning, \cite{Panda2016} trained a fully connected layer on the hidden representation of their encoder network for supervised learning, and report a 0.92\% classification error which is comparable to other state-of-the-art networks. In this work, hierarchical clustering was used instead of a classification of the latent representations, which again does not allow for a numerical comparison.

\paragraph{Conclusion}

Both \cite{Panda2016} and \cite{Burbank2015} follow the idea to further integrate computational neuroscience and machine learning. They independently develop different approaches for unsupervised feature learning from MNIST and CIFAR-10 images. Unfortunately, several relevant implementation details are unknown (e.g. discrete time step size, software used for implementation experiments). No code is available for either project, no comparison to non-spiking baseline models is conducted, no noise is added to the input data, and no processing times are available for comparison. 

The approach presented in this work is certainly more similar to the work by \cite{Panda2016} than \cite{Burbank2015}, based on the use of hierarchically stacked convolutional layers and error backpropagation. However, here I aimed to address some of the shortcomings of their approach. Notably, I succeeded in training multi-layer SAE networks end-to-end directly without complicated data pre-processing. Further, I compared network dynamics and performance to non-spiking baseline models trained on the same task, and improved transparency by using open-source software and making all code publicly available. 

\end{document}